\documentclass[manuscript,screen]{acmart}

\usepackage{multirow} 
\usepackage{array}

\usepackage{amsmath}
\usepackage{amssymb}
\usepackage{mathtools}
\usepackage{amsthm}
\usepackage{enumitem}

\usepackage{algorithm}
\usepackage{algpseudocode}

\usepackage{subfiles} %
\usepackage{makecell} %
\usepackage{multirow} %
\usepackage{subcaption}
\usepackage{color,xcolor}
\usepackage{graphicx}
\usepackage{svg}
\usepackage{booktabs}
\usepackage{makecell}
\usepackage{colortbl}
\usepackage{pifont}%
\usepackage{algorithm} %
\usepackage{listings}
\usepackage[switch]{lineno}
\usepackage{amssymb} %

\usepackage[misc]{ifsym}
\usepackage{colortbl}
\usepackage[normalem]{ulem}

\usepackage[capitalize,noabbrev]{cleveref}
\usepackage{tcolorbox}
\usepackage{hyperref}
\usepackage{longtable}
\usepackage{tikz}
\usetikzlibrary{positioning, arrows.meta}

\theoremstyle{plain}

\theoremstyle{definition}

\theoremstyle{remark}

\usepackage{float}

\usepackage[table]{xcolor} 
\usepackage{multirow}
\usepackage{hhline}        
\usepackage[textsize=tiny]{todonotes}
\usepackage{xspace}
\definecolor{shapecolor}{rgb}{0.0,0.5,0.0}

\definecolor{arylideyellow}{rgb}{0.91, 0.84, 0.42}

\usepackage[table]{xcolor}
\usepackage{colortbl}

\definecolor{lightgray}{RGB}{245, 245, 247} 

\definecolor{lightgreen}{RGB}{220, 245, 220}

\definecolor{lightorange}{RGB}{255, 235, 215}

\definecolor{lightred}{RGB}{255, 225, 225}

\newcommand{\benchName}{\texttt{WS-IMUBench}\xspace}
\AtBeginDocument{%
  }

\setcopyright{acmlicensed}
\copyrightyear{2018}
\acmBooktitle{Conference acronym 'XX}
\acmYear{2026}
\acmMonth{2}
\acmDOI{XXXXXXX.XXXXXXX}
\acmISBN{978-1-4503-XXXX-X/18/06}

\begin{document}

\title{
\benchName: Can Weakly Supervised Methods from Audio, Image, and Video Be Adapted for IMU-based Temporal Action Localization?}

\author{Pei Li}
\orcid{0009-0006-3194-556X}
\affiliation{
  \institution{School of Software Engineering, Xi'an Jiaotong University}
 \city{Xi'an}
 \state{Shaanxi}
 \country{China}}
 \email{lp@stu.xjtu.edu.cn}

\author{Jiaxi Yin}
\authornote{Co-author equal contribution.\\$\dagger$Corresponding author.  }
\orcid{0009-0009-5632-7890}
\affiliation{
 \institution{School of Software Engineering, Xi'an Jiaotong University}
 \city{Xi'an}
 \state{Shaanxi}
 \country{China}}
\email{jiaxiyin@stu.xjtu.edu.cn}

\author{Lei Ouyang}
\orcid{0009-0005-7941-6766}
\affiliation{
 \institution{School of Software Engineering, Xi'an Jiaotong University}
 \city{Xi'an}
 \state{Shaanxi}
 \country{China}}
\email{ouyanglei_20@stu.xjtu.edu.cn}

\author{Shihan Pan}
\orcid{0009-0006-9135-8922}
\affiliation{
 \institution{School of Software Engineering, Xi'an Jiaotong University}
 \city{Xi'an}
 \state{Shaanxi}
 \country{China}}
\email{16605986769@stu.xjtu.edu.cn}

\author{Ge Wang}
\orcid{0000-0002-9058-8543}
\affiliation{
 \institution{School of Computer Science and Technology, Xi'an Jiaotong University}
 \city{Xi'an}
 \state{Shaanxi}
 \country{China}}
\email{gewang@xjtu.edu.cn}

\author{Han Ding}
\orcid{0000-0002-5274-7988}
\affiliation{
 \institution{School of Computer Science and Technology, Xi'an Jiaotong University}
 \city{Xi'an}
 \state{Shaanxi}
 \country{China}}
\email{dinghan@xjtu.edu.cn}

\author{Fei Wang}
\authornotemark[2]
\orcid{0000-0002-0750-6990}
\affiliation{
 \institution{School of Software Engineering, Xi'an Jiaotong University}
 \city{Xi'an}
 \state{Shaanxi}
 \country{China}
 }
\email{feynmanw@xjtu.edu.cn}

\begin{abstract}
IMU-based Human Activity Recognition (HAR) has enabled a wide range of ubiquitous computing applications, yet its dominant “clip classification” paradigm cannot capture the rich temporal structure of real-world behaviors. This motivates a shift toward IMU Temporal Action Localization (IMU-TAL), which predicts both action categories and their start/end times in continuous streams. However, current progress is strongly bottlenecked by the need for dense, frame-level boundary annotations, which are costly and difficult to scale.  To address this bottleneck, we introduce \benchName, a systematic benchmark study of weakly supervised IMU-TAL (WS-IMU-TAL) under only sequence-level labels. Rather than proposing a new localization algorithm, we evaluate how well established weakly supervised localization paradigms from \emph{audio}, \emph{image}, and \emph{video} transfer to IMU-TAL under only sequence-level labels. We benchmark seven representative weakly supervised methods on seven public IMU datasets, resulting in over 3,540 model training runs and 7,080 inference evaluations. Guided by three research questions on transferability, effectiveness, and insights, our findings show that (i) transfer is modality-dependent, with temporal-domain methods generally more stable than image-derived proposal-based approaches; (ii) weak supervision can be competitive on favorable datasets (e.g., with longer actions and higher-dimensional sensing); and (iii) dominant failure modes arise from short actions, temporal ambiguity, and proposal quality. Finally, we outline concrete directions for advancing WS-IMU-TAL (e.g., IMU-specific proposal generation, boundary-aware objectives, and stronger temporal reasoning). Beyond individual results, \benchName establishes a reproducible benchmarking template, datasets, protocols, and analyses, to accelerate community-wide progress toward scalable WS-IMU-TAL.
\end{abstract}

\ccsdesc[500]{Human-centered computing~Ubiquitous and mobile computing systems and tools}

\keywords{temporal action localization, human action recognition, weakly supervised learning, wearable devices, IMU}

\received{20 February 2007}
\received[revised]{12 March 2009}
\received[accepted]{5 June 2009}


\maketitle

\section{Introduction}\label{sec:introduction}


Inertial Measurement Unit (IMU) based Human Activity Recognition (HAR) has become a foundational technology for pervasive computing, driving innovations in mobile health~\cite{subasi2018iot,bibbo2023human,gaud2024mhcnls,bhattacharya2022ensem,straczkiewicz2021systematic}, sports analytics~\cite{hoelzemann2023hang,mekruksavanich2022sport,ramasamy2018recent,pajak2022approach}, and industrial safety~\cite{suh2023wearable,makela2021introducing,sopidis2022micro,mekruksavanich2023automatic}. Traditionally, IMU-based HAR has been framed as a segment-based classification task, where a model identifies a single action label for a pre-segmented sensor snippet. However, this paradigm faces significant limitations in real-world deployments. First, it assumes activities are isolated, failing to capture the continuity of human behavior. Second, it lacks the granularity required for complex tasks such as behavioral pattern discovery, automated activity summarization, or fine-grained health diagnostics, where the exact onset and offset of actions are critical.

To address these shortcomings, the field is shifting toward Temporal Action Localization (TAL)~\cite{wang2025ego4o,bock2024wear,bock2024temporal,lan2025xrf}. Unlike simple classification, TAL aims to identify a sequence of actions within a continuous stream, providing both the action category and its precise temporal boundaries (start and end times). This provides a much richer understanding of user context. Recognizing these benefits, the community has recently introduced large-scale datasets such as WEAR~\cite{bock2024wear} and XRF-V2~\cite{lan2025xrf} to benchmark temporal localization.

Despite its potential, the progress of TAL is bottlenecked by the requirement for strong supervision. Training these models requires frame-by-frame annotations of action boundaries—a process that is notoriously costly, labor-intensive, and difficult to scale. Annotators must painstakingly scrub through hours of sensor data or synchronized video to mark precise timestamps, which is prone to human error and inter-annotator inconsistency.

In contrast, weakly supervised labeling offers a more scalable alternative. In this setting, the model is trained using only ``video-level" labels—a list of actions present in a long sensor sequence without any timing information. Such labels are significantly easier to obtain; for instance, users can simply recall the sequence of activities performed, or follow a pre-defined activity protocol where subjects execute a list of actions in a prescribed order without timestamp logging. Enabling TAL under weak supervision would not only reduce annotation overhead but also unlock the potential of massive, unlabelled wearable datasets.

To the best of our knowledge, Weakly Supervised Temporal Action Localization (WSTAL) specifically for IMU data remains an under-explored frontier. However, analogous challenges have been extensively studied in related domains, such as Weakly Supervised Video-based Action Localization (WSVAL)~\cite{lee2021weakly,zhang2021cola,shi2020weakly,lee2020background,nguyen2019weakly,shou2018autoloc}, Weakly Supervised Sound Event Detection (WSSED)~\cite{miyazaki2020weakly,lee2017ensemble,dinkel2021towards,xin2023improving}, and Weakly Supervised Object Detection (WSOD)~\cite{bilen2016weakly,tang2017multiple,tang2018pcl,ren2020instance}. These tasks share a fundamental structural requirement: the model must derive fine-grained localization (predicting temporal boundaries or spatial coordinates) from coarse-grained supervision (categorical presence-absence labels). Specifically, they all rely on the principle of discriminative localization, where the model learns to identify the most indicative segments or regions that contribute to the sequence-level classification.

\begin{figure}[t]
    \centering
    \includegraphics[width=1\linewidth]{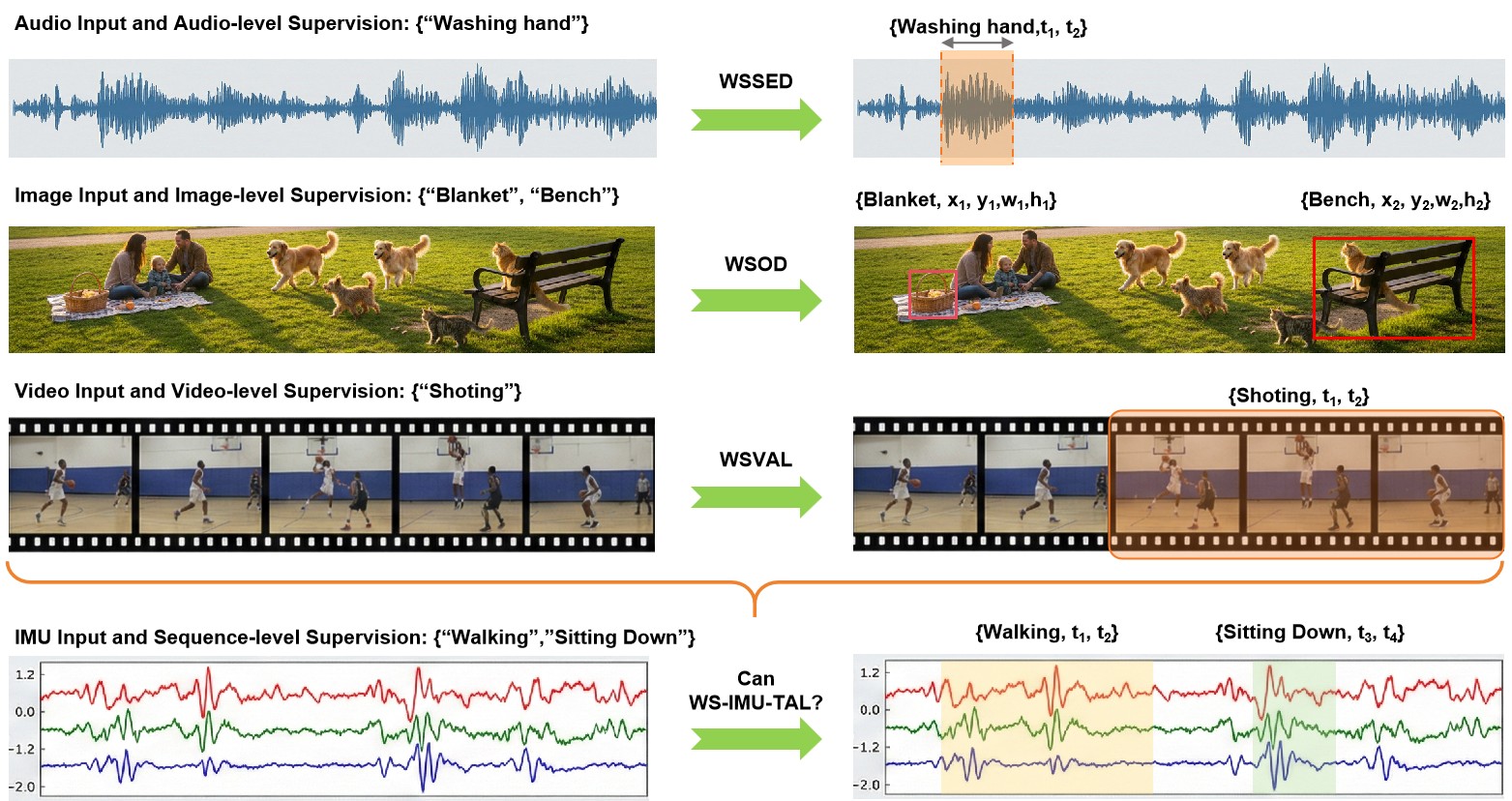}
    \caption{In established domains like weakly supervised sound event detection from audio (WSSED), weakly supervised object detection from images (WSOD), and weakly supervised video-based action localization (WSVAL), Weakly Supervised Learning has proven highly effective. These methods all use only coarse, audio/image/video-level labels to infer fine-grained boundaries (e.g., bounding boxes or temporal onset/offset). Can this successful paradigm be analogously applied to IMU data to achieve Weakly Supervised Temporal Action Localization (WS-IMU-TAL)?}
    \label{fig:weakly_story}
\end{figure}

In this work, we bridge this cross-modal gap by 
\emph{systematically} studying how weakly supervised localization mechanisms originally developed for identifying ``active'' regions in audio, video, and images---translate to continuous IMU streams with distinct characteristics (e.g., periodic motion patterns, sensor noise, subject variability, and ambiguous action boundaries). As illustrated in Figure~\ref{fig:weakly_story}, these domains have shown that coarse labels can still yield fine-grained localization, motivating our study of whether the same paradigm can be transferred to IMU-TAL.
Importantly, our contribution is not an incremental tweak of a single model; instead, we provide the missing \emph{foundational evidence and design guidance} needed to make weak supervision for IMU temporal localization a well-defined, reproducible, and comparable research problem. Specifically, we identify which components transfer, where naive transfer breaks, and what IMU-specific design choices are necessary for stable training and reliable boundary prediction.

To this end, we structure our investigation around the following three research questions:

\begin{itemize}
    \item \textbf{RQ1 (Transferability):} How can weakly supervised learning paradigms that have proven effective in other modalities (e.g., Multiple Instance Learning based localization) be adapted to continuous IMU streams? What IMU-specific design choices are required to define instances and apply coarse supervision for temporal localization?
    \item \textbf{RQ2 (Effectiveness):} To what extent do these transferred weakly supervised approaches enable accurate recognition and temporal localization on IMU data? How do they perform across datasets with diverse sensing conditions and activity characteristics?
    \item \textbf{RQ3 (Insights \& Future Directions):} What are the dominant failure modes and limiting factors of weakly supervised IMU temporal localization, and what opportunities do they reveal for advancing WS-IMU-TAL, such as better temporal reasoning and robustness to action overlaps?
\end{itemize}

\paragraph{Findings (Summary).}
Our study yields three key findings.
\begin{itemize}
    \item \textbf{(F1) Transferability is partial and modality-dependent.} Temporal-domain methods transferred from audio/video (e.g., WSSED and WSTAL) are generally more stable on IMU streams than image-derived WSOD methods, largely due to better alignment with the sequential structure of inertial signals.
    \item \textbf{(F2) Weak supervision can be competitive on favorable datasets.} Without frame-level boundary annotations, weakly supervised models achieve meaningful localization, and can narrow the gap to fully supervised TAL on datasets with longer action durations and higher-dimensional sensor inputs (e.g., RWHAR and WEAR).
    \item \textbf{(F3) The dominant bottlenecks are proposals, short actions, and temporal ambiguity.} Performance degrades sharply on datasets dominated by very short actions or low-dimensional sensing; for proposal-based WSOD transfers, imperfect temporal proposals further cap the achievable localization accuracy.
\end{itemize}

\paragraph{Contributions.}
We position WS-IMU-TAL as a distinct and practically important problem, and contribute a rigorous and reproducible benchmark study that goes beyond simply running more baselines. Concretely, our contributions focus on \emph{standardization}, \emph{diagnostic evaluation}, and \emph{actionable insights}:

\begin{itemize}
    \item \textbf{Problem formulation + standardized benchmark setup.} We formalize WS-IMU-TAL under sequence-level supervision and standardize the task setup (datasets, splits, and inference modes) to make results comparable across methods.
    \item \textbf{Unified evaluation protocol beyond mAP.} We establish an evaluation protocol that jointly measures recognition and temporal localization via complementary metrics (frame-level labeling metrics, misalignment ratios, and segment-level mAP), enabling fine-grained error diagnosis rather than a single aggregated score.
    \item \textbf{Cross-modal method adaptation with IMU-specific design choices.} We adapt representative weakly supervised localization paradigms from audio, image, and video to IMU streams, and distill the IMU-specific choices (e.g., instance definition, bag construction, temporal aggregation, and proposal generation) that critically affect stability and performance.
    \item \textbf{Large-scale empirical study + transferable takeaways.} We conduct extensive experiments across seven public IMU datasets and analyze dominant failure modes (e.g., short actions, temporal ambiguity, and proposal quality), yielding practical guidance for future WS-IMU-TAL research.
\end{itemize}

\section{Related Work}\label{sec:related_work}

\subsection{IMU-based Human Action Recognition and Action Localization}\label{sec:imu-har}

Traditionally, IMU-based Human Activity Recognition (HAR) has been formulated as window-level classification on pre-segmented sliding windows. Early deep models, such as DeepConvLSTM by Ord\'{o}\~{n}ez and Roggen~\cite{ordonez2016deep}, combined convolutional and recurrent layers to learn spatiotemporal representations from raw sensor streams. Subsequent work improved discriminative learning via attention and metric-style regularization, e.g., the ``Attend and Discriminate" framework~\cite{abedin2021attend}, while deployment-oriented studies proposed lightweight architectures such as TinyHAR~\cite{zhou2022tinyhar}. Despite strong performance on curated benchmarks, window-based HAR fundamentally assumes that activities are already segmented into isolated clips and can be represented at a fixed temporal scale. This assumption breaks down in real-world, continuous sensing, where actions occur sequentially and transitions are not aligned with window boundaries. Moreover, window-level classification provides only coarse labels and lacks the temporal granularity required by downstream applications such as behavioral pattern discovery, activity summarization, and fine-grained health monitoring, where accurate onset/offset estimation is essential.

To overcome these limitations, recent studies have started to adapt Temporal Action Localization (TAL) paradigms from computer vision~\cite{xia2020survey} to the IMU domain. Bock et al.~\cite{bock2024temporal} provided a systematic investigation of transferring state-of-the-art video TAL models to inertial signals, showing that single-stage detectors such as ActionFormer~\cite{zhang2022actionformer}, TriDet~\cite{shi2023tridet}, and TemporalMaxer~\cite{tang2023temporalmaxer} can be effectively trained on IMU datasets. These models leverage long-range temporal modeling (e.g., attention-based feature aggregation and scalable temporal perception) and substantially outperform sliding-window baselines for localization.

However, existing IMU-TAL research is still dominated by fully supervised training, which requires precise temporal boundaries for each action instance. In contrast to visual or auditory recordings, IMU streams are abstract multi-channel waveforms without intuitive semantics, making manual boundary annotation difficult, time-consuming, and prone to inconsistency~\cite{bock2024wear}. This high annotation cost motivates exploring weakly supervised IMU-TAL, where only sequence-level labels are available while the model must infer action boundaries.

\subsection{Weakly Supervised Learning}\label{sec:wsl}

Weakly supervised learning (WSL) seeks to learn instance-level recognition and/or localization from coarse supervision, such as image-level or clip-level labels, when fine-grained annotations are expensive or unavailable.
In this section, we organize prior WSL studies by \textbf{audio}, \textbf{image}, and \textbf{video} modalities, because our goal is to examine how representative weakly supervised localization mechanisms developed in these domains can be adapted to IMU-based TAL.

\paragraph{Weakly supervised learning in audio.}

In the audio domain, weakly supervised learning primarily targets Sound Event Detection (SED) using time-frequency inputs (e.g., spectrograms)~\cite{kong2020panns}. As sound signals are 1D time-series data, they share intrinsic physical properties---such as signal continuity and temporal dependency---with IMU sensor readings. The Convolutional Recurrent Neural Network (CRNN) has emerged as the dominant architecture for modeling these sequences, serving as the standard baseline in DCASE benchmarks~\cite{cakir2017convolutional, turpault2019sound}. Although early approaches relied on global maximum or average pooling, recent research has shifted towards learnable aggregation strategies (often formulated as MIL-style pooling) to mitigate localization ambiguity. Prominent methods include Attention Pooling (ATP)~\cite{wang2019comparison, ilse2018attention} and adaptive pooling~\cite{mcfee2018adaptive}, which dynamically weight informative frames and provide a proxy for frame-level localization. Furthermore, to address the instability of predicting events at varying lengths, CDur~\cite{dinkel2021towards} incorporates duration robustness through specialized linear softmax pooling and consistency regularization.


\paragraph{Weak supervision in images.}
In the image domain, weakly supervised learning is most relevant to our setting through \emph{Weakly Supervised Object Detection (WSOD)}, where only image-level labels are available but the goal is to localize instances. A common formulation is Multiple Instance Learning (MIL), treating an image as a bag of region proposals and learning proposal-level scores from bag-level supervision~\cite{bilen2016weakly}. Representative WSOD pipelines such as WSDDN~\cite{bilen2016weakly} use a dual-stream scoring mechanism to couple classification and localization, while follow-up works improve localization via iterative refinement (OICR~\cite{tang2017multiple}) and more stable pseudo-label generation (PCL~\cite{tang2018pcl}). These methods highlight recurring challenges under weak supervision, such as focusing on only the most discriminative parts and drifting pseudo labels across iterations.


\paragraph{Weak supervision in videos.}

Weakly-supervised temporal action localization (WSTAL) aims to identify action segments with only video-level labels, a task commonly framed under the Multiple Instance Learning (MIL) paradigm~\cite{nguyen2018weakly,ma2021weakly}. Early efforts focused on improving the selection of discriminative snippets within the MIL framework. For instance, STPN \cite{nguyen2018weakly} introduced a sparsity constraint to pinpoint key moments, while AutoLoc \cite{shou2018autoloc} designed a specific loss function to encourage more precise temporal boundaries. Beyond snippet selection, representative directions include contrastive feature refinement (e.g., CoLA~\cite{zhang2021cola}) and explicit background/context modeling~\cite{nguyen2019weakly,zhang2019glnet} to avoid collapsing onto isolated peaks. Some studies further incorporate sparse point-level cues to guide boundary learning~\cite{lee2021learning}.


\paragraph{Summary and gap.}
Although weak supervision has been extensively studied in audio, image, and video domains, its systematic adoption for IMU-based TAL remains limited. In particular, prior IMU-TAL studies mainly focus on full supervision, leaving open questions on (i) how to reliably translate MIL-style bag/instance designs and aggregation mechanisms to IMU streams, (ii) how effective these transferred approaches are across diverse IMU benchmarks, and (iii) what dominant failure modes emerge under weak supervision.
This paper addresses these gaps by providing a unified cross-modal adaptation and evaluation of representative WSL mechanisms, together with an analysis of their limitations and opportunities for future WS-IMU-TAL research.

\section{Weakly Supervised Learning for IMU Data}\label{sec:weakly-tal}

\paragraph{\textbf{Answering RQ1 (Transferability).}}
RQ1 asks how weakly supervised localization paradigms from other modalities can be adapted to continuous IMU streams.
In this section, we answer RQ1 by casting WS-IMU-TAL as a Multiple Instance Learning (MIL) problem. We then show how three representative lines of prior work, i.e., slice-level MIL pooling from audio, proposal-based MIL from images, and snippet-based WSTAL from videos, can be transferred by re-defining the instance space on the temporal axis and by adopting IMU-specific choices for windowing, aggregation, and temporal refinement.

\subsection{Problem Definition}
\label{sec:problem-definition}

\paragraph{\textbf{(1) HAR, TAL, and WSTAL}}
Human activity analysis from sensor data encompasses a spectrum of tasks, primarily Human Activity Recognition (HAR) and Temporal Action Localization (TAL). Let an untrimmed multivariate sequence be $X \in \mathbb{R}^{S \times T_{\mathrm{raw}}}$, where $S$ denotes the number of sensor channels and $T_{\mathrm{raw}}$ denotes the length of the raw sequence (number of time steps), and let $\mathcal{C}=\{1,\dots,C\}$ be the set of activity classes. The distinction between these tasks lies in their objectives and the nature of their supervision. 

\begin{itemize}
    \item Supervised HAR : the goal is to predict a single label $a \in \mathcal{C}$ for a pre-segmented, short clip $x \subset X$. The training data consists of pairs $(x, a)$, making it a standard classification task~\cite{anguita2012human, heydarian2023rwisdm}.

    \item Supervised TAL : this task operates on the entire sequence $X$ under the supervision of a set of annotations $\mathcal{Y}=\{(c_i, s_i, e_i)\}_{i=1}^{N}$, where each tuple contains the class label $c_i \in \mathcal{C}$ and precise start/end boundaries $(s_i, e_i)$.  The goal of Supervised TAL is to produce a set of predicted actions with corresponding start/end boundaries.

    \item Weakly Supervised TAL (WSTAL): the goal remains the same as the supervised TAL. However, the available supervision is drastically reduced. For each sequence $X$, we are only given a single multi-hot action label $\mathbf{y} \in \{0,1\}^C$, which merely indicates the presence or absence of each action class. The ground-truth start/end boundaries in annotations $\mathcal{Y}$ is not provided during training.
\end{itemize}

\paragraph{\textbf{(2) The Multiple Instance Learning Paradigm}}

In the weakly supervised setting above, each untrimmed stream $X$ is annotated only with a coarse, sequence-level multi-hot label $\mathbf{y}$, without any temporal boundary information.
This form of coarse supervision is commonly referred to as \textbf{bag-level supervision} in the MIL literature.
Accordingly, this problem is canonically addressed by the MIL framework.
In MIL, a single untrimmed sequence is treated as a \textbf{bag}.
We explicitly define the key concepts as follows:
\begin{itemize}
    \item \textbf{Bag:} one long, untrimmed recording (here, an IMU stream $X$).
    \item \textbf{Instance:} a candidate temporal segment extracted from the bag (e.g., a sliding window or a proposal), denoted as $\{x_\ell\}_{\ell=1}^L$.
    \item \textbf{Bag label:} a coarse multi-hot label $\mathbf{y}$ indicating which classes appear in the bag (without temporal boundaries).
    \item \textbf{Positive instance assumption:} if $y_c=1$, the bag is assumed to contain at least one instance that truly belongs to class $c$ (while most other instances may be background).
\end{itemize}

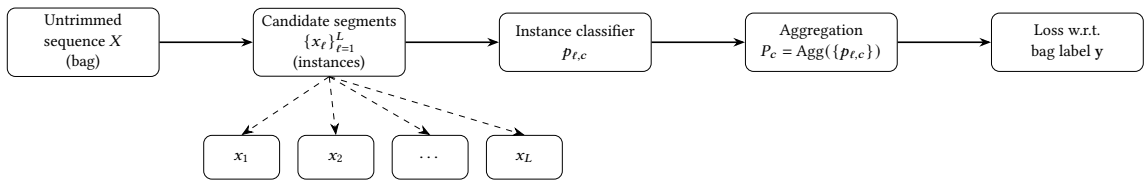
\begin{figure}[h]
    \centering
    \resizebox{\linewidth}{!}{%
    \begin{tikzpicture}[
        font=\small,
        box/.style={draw, rounded corners, align=center, minimum height=1.0cm, minimum width=2.6cm},
        smallbox/.style={draw, rounded corners, align=center, minimum height=0.75cm, minimum width=1.3cm},
        arr/.style={-{Stealth[length=2mm]}, thick}
    ]
        \node[box] (bag) {Untrimmed\\sequence $X$\\(bag)};
        \node[box, right=1.6cm of bag] (inst) {Candidate segments\\$\{x_\ell\}_{\ell=1}^{L}$\\(instances)};
        \node[box, right=1.6cm of inst] (score) {Instance classifier\\$p_{\ell,c}$};
        \node[box, right=1.6cm of score] (agg) {Aggregation\\$P_c=\text{Agg}(\{p_{\ell,c}\})$}; 
        \node[box, right=1.6cm of agg] (loss) {Loss w.r.t.\\bag label $\mathbf{y}$};

        \draw[arr] (bag) -- (inst);
        \draw[arr] (inst) -- (score);
        \draw[arr] (score) -- (agg);
        \draw[arr] (agg) -- (loss); 

        \node[smallbox, below=1.0cm of inst, xshift=-1.5cm] (x1) {$x_1$};
        \node[smallbox, right=0.3cm of x1] (x2) {$x_2$};
        \node[smallbox, right=0.3cm of x2] (xdots) {$\dots$};
        \node[smallbox, right=0.3cm of xdots] (xL) {$x_L$};
        
        \draw[arr, dashed, thin] (inst.south) -- (x1.north);
        \draw[arr, dashed, thin] (inst.south) -- (x2.north);
        \draw[arr, dashed, thin] (inst.south) -- (xdots.north);
        \draw[arr, dashed, thin] (inst.south) -- (xL.north);
    \end{tikzpicture}%
    }
    \caption{An overview of the Multiple Instance Learning (MIL) pipeline. Block diagram of a MIL pipeline: an untrimmed sequence (bag) is decomposed into candidate segments (instances), scored by an instance classifier, aggregated into a bag-level prediction, and trained with a loss against the bag label.}
    \label{fig:mil_overview}
\end{figure}

Under this assumption, the model generates candidate instances $\{x_\ell\}_{\ell=1}^L$, predicts per-instance probabilities $\{p_{\ell,c}\}_{\ell=1}^L$, and aggregates them via a permutation-invariant operator $\mathrm{Agg}(\cdot)$ (e.g., max or attention pooling) to obtain a bag-level prediction $P_c = \mathrm{Agg}(\{p_{\ell,c}\})$, as shown in Fig.~\ref{fig:mil_overview}. The model is trained using a standard classification loss between $P_c$ and the bag label $\mathbf{y}$.

\paragraph{\textbf{(3) MIL in Audio, Image, and Video}}
The bag/instance view in MIL is generic, but its concrete meaning depends on the data modality, i.e., what constitutes a bag, how instances are generated, and what aggregation best connects instance scores to the bag label.
Prior work has demonstrated that this abstraction transfers well across modalities by choosing appropriate instance definitions.
For sound event detection, a sequence (bag) is decomposed into short time-frequency patches or slices (instances), and MIL-style pooling aggregates slice-level predictions to match sequence-level labels~\cite{dinkel2021towards, cakir2017convolutional}. For images, an image (bag) contains multiple region proposals (instances), whose scores are aggregated to perform weakly supervised object detection~\cite{bilen2016weakly,tang2017multiple}. For videos, an untrimmed video (bag) is segmented into temporal snippets (instances) for weakly supervised temporal action localization~\cite{nguyen2018weakly,shou2018autoloc}. Across these domains, the core objective is the same: learn to localize fine-grained instances (temporal segments or spatial regions) using only coarse, bag-level supervision.


\paragraph{\textbf{(4) Analogy to IMU Data: Justifying the Transfer.}}
The key reason cross-modal transfer is feasible is that our weakly supervised IMU setting can be cast in the same MIL form as prior work: an untrimmed IMU stream is a \emph{bag} with only sequence-level labels, and the model must infer which temporal \emph{instances} (segments) explain the bag label.
Therefore, what we transfer from other modalities is not the raw input representation, but the \emph{bag/instance design} and the \emph{aggregation/localization mechanisms} that turn instance scores into bag predictions and localization cues.

\begin{itemize}
    \item \textbf{Analogy to Audio (slice-level MIL pooling):} Both IMU and audio are multi-channel 1D sequences. In weakly supervised sound event detection, a recording is decomposed into short slices (instances), and attention/soft pooling selects salient frames as evidence for the sequence (bag) label. Similarly, an IMU stream can be decomposed into short temporal units, where high-response units indicate potential action intervals.

    \item \textbf{Analogy to Images (proposal-based MIL):} In weakly supervised object detection, an image (bag) is represented by a set of candidate region proposals (instances), typically parameterized by a 2D bounding box $(x,y,w,h)$. Each proposal is scored and then aggregated to match the image-level label, while refinement mechanisms iteratively improve proposal assignments. For IMU, we can analogously represent an untrimmed sequence (bag) by a set of candidate temporal proposals with varying durations. Under this mapping, a 2D box reduces to a 1D temporal interval $(s,e)$, and WSOD-style proposal scoring and refinement can be reused after adapting proposal generation (e.g., sampling temporal segments) and incorporating basic temporal consistency.

    \item \textbf{Analogy to Videos (snippet-level WSTAL):} This is the most direct analogy. In video WSTAL, an untrimmed video (bag) is decomposed into a sequence of short snippets (instances), from which a feature encoder produces snippet-level representations and class activation scores that are aggregated by MIL to match video-level labels. IMU streams admit an analogous snippet view by treating short temporal clips (or their features) as instances in the bag. This allows us to transfer key WSTAL ingredients---snippet feature extraction, MIL-based aggregation, and temporal modeling modules (e.g., contrastive feature refinement and temporal consistency regularization)---after adapting them to IMU-specific sampling rates and noise.

\end{itemize}

At the same time, IMU signals introduce modality-specific challenges (e.g., sampling-rate variability, sensor noise, periodic motion, and ambiguous boundaries), which motivate careful instance definition and temporal aggregation design in our WS-IMU-TAL formulation.

\paragraph{\textbf{(5) Formal Definition for WS-IMU-TAL.}}  Leveraging the analogies above, we define Weakly Supervised IMU Temporal Action Localization (WS-IMU-TAL) as follows.
Given a multivariate IMU stream $\mathbf{x} \in \mathbb{R}^{S \times T}$ with $S$ sensor channels and $T$ time steps, and a bag-level multi-hot label vector $\mathbf{y} \in \{0,1\}^C$ indicating which of the $C$ action classes occur in the stream (without any temporal boundary annotations), the goal is to learn a model $f$ that outputs a (potentially empty) set of predicted action instances
\begin{equation}
\hat{\mathcal{Y}} = \{(\hat{c}_i, \hat{s}_i, \hat{e}_i)\}_{i=1}^{\hat{N}},
\end{equation}
where $\hat{N}$ is the number of predicted instances, $\hat{c}_i \in \{1,\dots,C\}$ is the predicted class, and $0 \le \hat{s}_i < \hat{e}_i \le T$ are the predicted start/end timestamps. 
During training, only $\mathbf{y}$ supervises $f$ via MIL-style aggregation over latent temporal instances; at inference, $f$ must output temporally precise segments whose union provides evidence for the bag-level label $\mathbf{y}$.

\subsection{Weakly Supervised TAL Architectures Overview}
\label{sec:architecture-overview}

To better understand the design principles and transferable components of weakly supervised temporal action localization (TAL), we review representative architectures developed in three different modalities: audio, images, and videos.

\paragraph{ \textbf{(1) Weakly Supervised TAL Architectures from audio}} Given the strong analogy between IMU and audio as high-frequency, multi-channel 1D sequences, we draw architectural inspiration from weakly supervised sound event detection (WSSED)~\cite{kong2020panns, wang2019comparison}.
Concretely, we adopt two representative lines of work: the DCASE baseline~\cite{turpault2019sound} for slice-level prediction with MIL aggregation, and CDur~\cite{dinkel2021towards} for regularizing temporal extent under weak supervision.

\begin{figure}[t]
    \centering
    \includegraphics[width=1\linewidth]{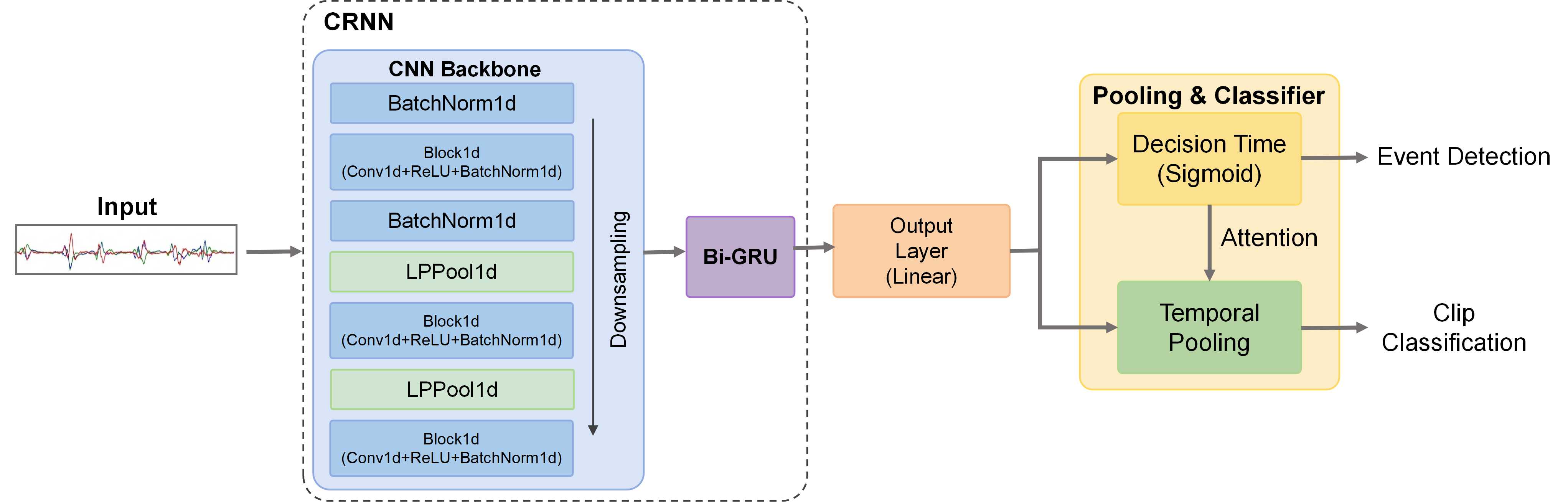}
    \caption{Network diagram of the CDur model: CNN feature extractor followed by a BiGRU for temporal modeling, with two output branches producing sequence-level predictions and upsampled slice-level predictions.}
    \label{fig:cdur_model}
\end{figure}

\textbf{Core Framework.}
We follow the standard WSSED recipe: a slice-level encoder produces temporally aligned predictions, MIL pooling bridges slice-level scores to sequence-level supervision, and an additional duration prior regularizes the temporal extent. We use a Convolutional Recurrent Neural Network (CRNN)~\ref{fig:cdur_model}, a standard backbone in WSSED, to encode an IMU stream into temporally aligned features. The CNN layers capture local motion patterns, while the Bidirectional GRU (BiGRU) models longer-range temporal dependencies. This yields slice-level class probabilities $\mathbf{P} = \{p_t \in \mathbb{R}^C\}_{t=1}^T$.

Since training labels are only sequence-level, we aggregate slice-level predictions into a bag-level prediction using attention-based MIL pooling. An attention branch produces weights $\alpha_t$ to emphasize informative frames, and the sequence-level prediction is computed as
\begin{equation}
\hat{\mathbf{y}} = \frac{\sum_{t=1}^T \alpha_t \cdot p_t}{\sum_{t=1}^T \alpha_t}.
\end{equation}
Importantly, the attention weights $\alpha_t$ also serve as a saliency signal for temporal localization, providing a high-resolution estimate of where actions likely occur.

A common failure mode of MIL-based localization is fragmented activations (over-segmentation) or missing non-salient portions (incomplete activation). CDur addresses this by introducing duration-aware constraints that encourage predicted activations to have realistic temporal spans.

\textbf{Adaptations for Sequential IMU Data.}
We adapt the WSSED framework to IMU by (1) treating short IMU clips as the basic temporal instances, (2) using a CRNN to encode multi-channel inertial signals (accelerometer/gyroscope axes) into slice-level action probabilities, (3) applying attention-based MIL pooling to match sequence-level multi-hot labels, and (4) incorporating a duration-based regularization term to reduce spurious short activations and to encourage temporally coherent segments.
This design directly transfers the ``frame scoring + MIL pooling + extent regularization'' pattern from audio to IMU, while remaining agnostic to the exact sampling rate and sensor placement.

\paragraph{ \textbf{(2) Weakly Supervised TAL Architectures from images}} We draw inspiration from weakly supervised object detection~ (WSOD)~\cite{bilen2016weakly,tang2018pcl,tang2017multiple}, where the goal is to localize objects using only image-level labels.
In our setting, we treat an untrimmed IMU stream as a bag and candidate temporal segments as proposals.

\begin{figure}[h]
    \centering
    \includegraphics[width=1\linewidth]{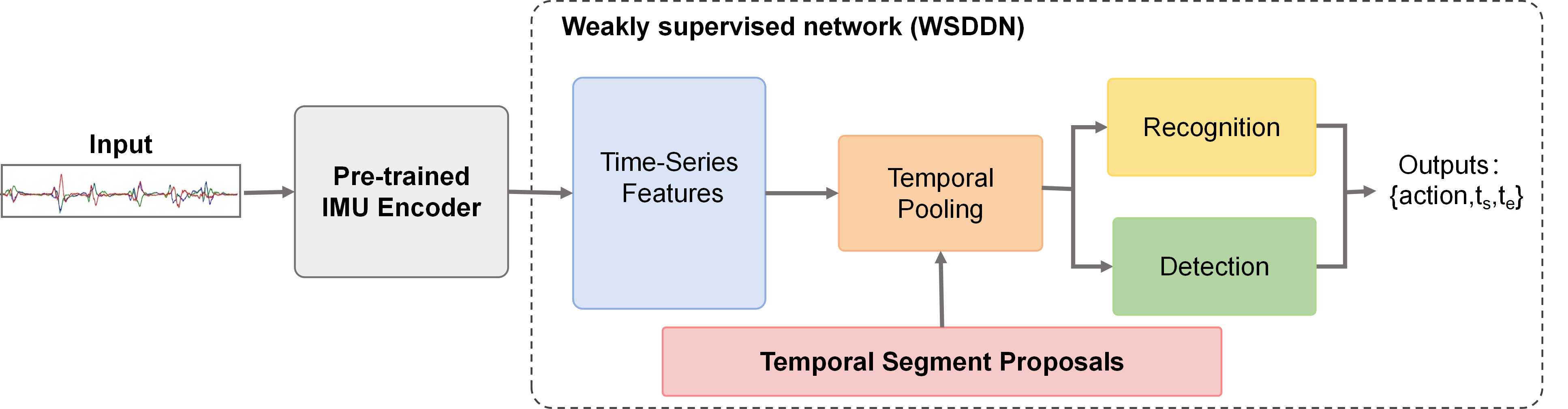}
    \caption{Our WSDDN-adapted architecture for IMU action localization. After feature extraction, each temporal proposal is processed by two parallel branches: a classification stream for class scores ($\mathbf{c}_p$) and a detection stream for localization weights ($\mathbf{l}_p$). The final score for each proposal is their element-wise product, and the sum of all scores is supervised by the sequence-level bag label.}
    \label{fig:wsddn_model}
\end{figure}

\textbf{Core Framework.}
Our base model, illustrated in Figure~\ref{fig:wsddn_model}, is adapted from WSDDN~\cite{bilen2016weakly}, which employs a dual-branch structure to compute classification scores $\mathbf{c}_p \in \mathbb{R}^{C}$ and localization weights $\mathbf{l}_p \in \mathbb{R}^{C}$ for each temporal proposal $r_p$. The final proposal score is an element-wise product:
\begin{equation}
    \mathbf{s}_p = \mathbf{c}_p \odot \mathbf{l}_p.
\end{equation}
The bag-level prediction is then obtained by summing over all proposals, $\hat{\mathbf{y}} = \sum_{p=1}^{P} \mathbf{s}_p$, and is supervised by a sample-level multi-label classification loss.
To further enhance localization accuracy, we extend this base model by incorporating the multi-stage refinement strategies from OICR~\cite{tang2017multiple} and PCL~\cite{tang2018pcl}. OICR's strategy of using high-scoring proposals to iteratively supervise subsequent stages, and PCL's approach of clustering proposals to generate stable pseudo-labels, are both adapted to operate on top of the proposal scores generated by our base model.

\begin{figure}[t]
    \centering
    \includegraphics[width=0.8\linewidth]{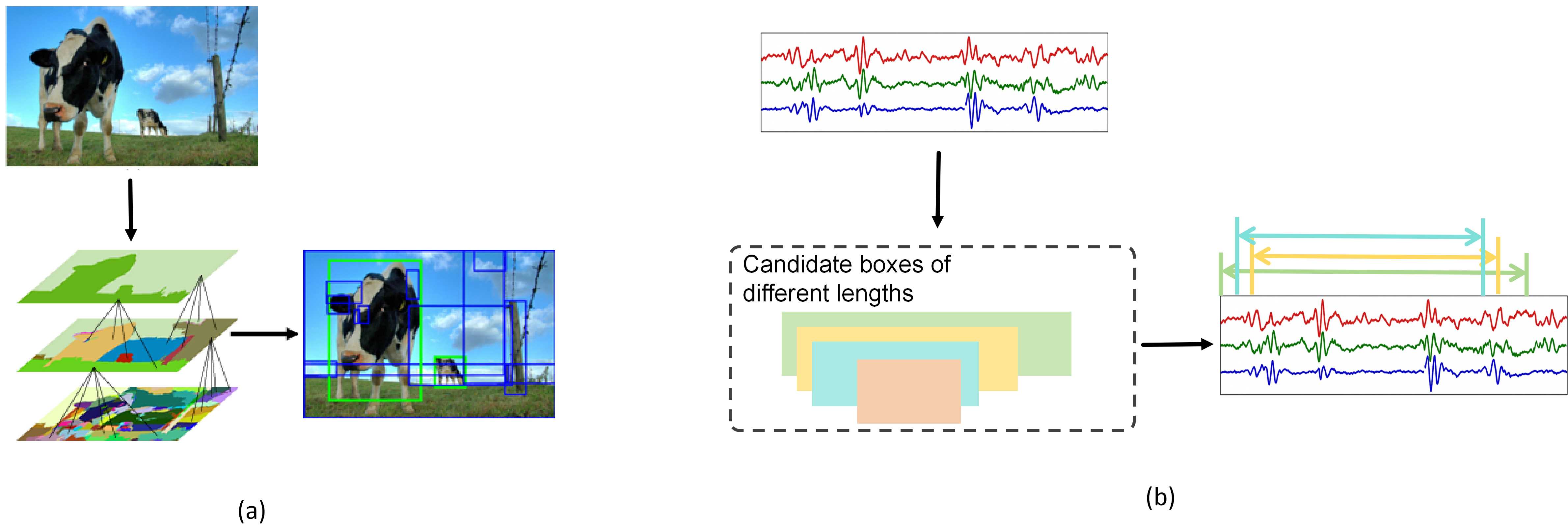}
    \caption{Comparison of proposal generation strategies. \textbf{(a) Selective Search for Images:} This algorithm leverages rich visual cues (color, texture) to intelligently propose a sparse set of candidate object regions, avoiding an inefficient brute-force search. \textbf{(b) Our Multi-Scale Temporal Sampling for IMU Data:} Lacking visual cues, our method adapts this concept to the time domain. It systematically samples temporal segments across multiple predefined durations to generate a diverse set of candidate instances, serving as an efficient, content-agnostic alternative for 1D signals. (see Algorithm~\ref{alg:proposal_boxes}.)}
    \label{fig:proposal_generations}
\end{figure}

\textbf{Adaptations for Sequential IMU Data.}
To successfully apply the image-based WSOD framework to our task, we apply the following minimal adaptations for 1D sequential data:
(1) Paradigm Analogy: We translate the ``bag-of-instances" concept from images to the time domain. A long, untrimmed IMU sequence is treated as the ``bag," and its temporal segments are considered the ``instances."
(2) Pretrained Feature Extractor: Analogous to a CNN backbone pretrained on a large-scale dataset like ImageNet~\cite{deng2009imagenet}, we employ a feature extractor $g(\cdot)$ that is pretrained on a supervised action classification task using short IMU clips. This extractor encodes the input sequence $\mathbf{x} \in \mathbb{R}^{S \times T}$ into a series of feature vectors $\Phi=\{\phi_\ell \in \mathbb{R}^d\}_{\ell=1}^L$.
(3) Temporal Proposal Generation: As image-based proposal algorithms like Selective Search~\cite{van2011segmentation} are inapplicable to 1D signals, we designed a hybrid sampling algorithm to generate candidate temporal segments (see Figure~\ref{fig:proposal_generations} and Algorithm~\ref{alg:proposal_boxes}). To ensure proposals are plausible, their duration range is first informed by the statistics of ground-truth actions for each specific dataset, and is capped by the input length for window-based analysis. We then generate a standardized quantity of 3000 proposals per sequence larger than the 2000 typical for Selective Search~\cite{van2011segmentation} to provide a rich candidate pool and ensure fair model comparisons. This is achieved via a hybrid strategy: a structured sliding-window approach generates the majority of proposals (70\%), while fully random sampling provides the remaining 30\% to enhance diversity. This process yields a diverse final set of $R=3000$ temporal segments, $\{r_i\}_{i=1}^{R}$, which are then used to pool features to serve as the instances for our MIL model.

\begin{algorithm}[ht]
\caption{Generate Proposal Boxes}
\label{alg:proposal_boxes}
\begin{algorithmic}[1]
\State \textbf{Input:} $T_{\text{global}}, N, \text{fps}, W, \text{min\_sec}, \text{max\_sec}, \text{sec\_resolution}, \text{fixed\_keep\_ratio}$
\State \textbf{Output:} Tensor $P \in \mathbb{R}^{N,2}$

\State $r \gets \text{Random}(seed)$
\If{$\text{raw\_frames}$ is None}
    \State $\text{raw\_frames} \gets \text{round}(W \cdot \text{fps})$
\EndIf

\State $r_f \gets \frac{T_{\text{global}}}{\text{raw\_frames}}$
\State $S \gets \{ s=s_{\text{min}} + i \cdot \text{sec\_resolution} \mid s_{\text{min}} \leq s \leq s_{\text{max}} \}$

\For{each $s \in S$}
    \State $\text{feat\_len} \gets \text{round}(s \cdot \text{fps} \cdot r_f)$
    \State $boxes \gets \left[ [s, s + \text{feat\_len}] \mid s + \text{feat\_len} \leq T_{\text{global}} \right]$
    \State $pool[s] \gets {boxes}$
\EndFor

\If{$\text{len}(pool) = 0$}
    \State \Return $P \gets [[0, T_{\text{global}}]] \times N$
\EndIf

\State $N_{\text{fixed}} \gets \text{round}(N \cdot \text{fixed\_keep\_ratio})$
\State $P_{\text{fixed}} \gets []$
\For{each $s \in S$}
    \State $q[s] \gets \text{min}( \text{fixed\_per\_scale\_min}, \text{len}(pool[s]))$
    \State $P_{\text{fixed}} \gets P_{\text{fixed}} \cup \text{sample}(pool[s], q[s])$
\EndFor

\While{$\text{len}(P) < N$}
    \State $\text{dur} \gets r.\text{uniform}(s_{\text{min}}, s_{\text{max}})$
    \State $\text{start\_sec} \gets r.\text{uniform}(0, W - \text{dur})$
    \State $\text{feat\_start} \gets \text{round}(\text{start\_sec} \cdot \text{fps} \cdot r_f)$
    \State $P \gets P \cup [\text{feat\_start}, \text{feat\_start} + \text{dur} \cdot \text{fps} \cdot r_f]$
\EndWhile

\State \Return $P$
\end{algorithmic}
\end{algorithm}

\paragraph{\textbf{(3) Weakly Supervised TAL Architectures from videos.}} 
Weakly supervised video-based action localization (WSVAL) is the closest counterpart to our setting: an untrimmed sequence (bag) is decomposed into snippets (instances), trained only with video-level labels, and the model must output start/end boundaries.
A long-standing challenge in WSVAL is \emph{incomplete localization}, where MIL training encourages the model to focus only on the most discriminative snippets instead of the full temporal extent of an action~\cite{shou2018autoloc,nguyen2018weakly,zhang2021cola}.

\textbf{Core Framework}. A typical WSVAL model consists of: (1) a snippet feature encoder $g(\cdot)$ (e.g., a temporal CNN/Transformer) that maps the untrimmed sequence to features $\Phi=\{\phi_t\}_{t=1}^{T}$; (2) a classifier producing class activation scores (CAS) $\mathbf{s}_t\in\mathbb{R}^{C}$; and (3) a permutation-invariant MIL aggregator (e.g., top-$k$ pooling or attention) that converts snippet scores to video-level predictions for multi-label supervision. 
Localization is then obtained by thresholding CAS (often after smoothing) and applying NMS/merging to generate segments.

CoLA~\cite{zhang2021cola} improves snippet representations by adding a contrastive objective that pulls together features belonging to the same latent action while pushing apart background/other-action snippets.
For an anchor snippet feature $\phi_i$ and a positive counterpart $\phi_p$ from the same action instance, the loss maximizes their similarity and minimizes similarity to $N$ negatives $\{\phi_n\}_{n=1}^N$:
\begin{equation}
    \mathcal{L}_{\text{CoLA}} = -\log \frac{\exp(\text{sim}(\phi_i, \phi_p) / \tau)}{\exp(\text{sim}(\phi_i, \phi_p) / \tau) + \sum_{n=1}^{N} \exp(\text{sim}(\phi_i, \phi_n) / \tau)},
\end{equation}
where $\text{sim}(\cdot,\cdot)$ is cosine similarity and $\tau$ is a temperature hyperparameter.
Intuitively, by making different parts of the same action more feature-consistent, subsequent MIL pooling becomes less biased toward only the peak snippets, which improves temporal coverage.

\begin{figure}[t]
    \centering
    \includegraphics[width=1\linewidth]{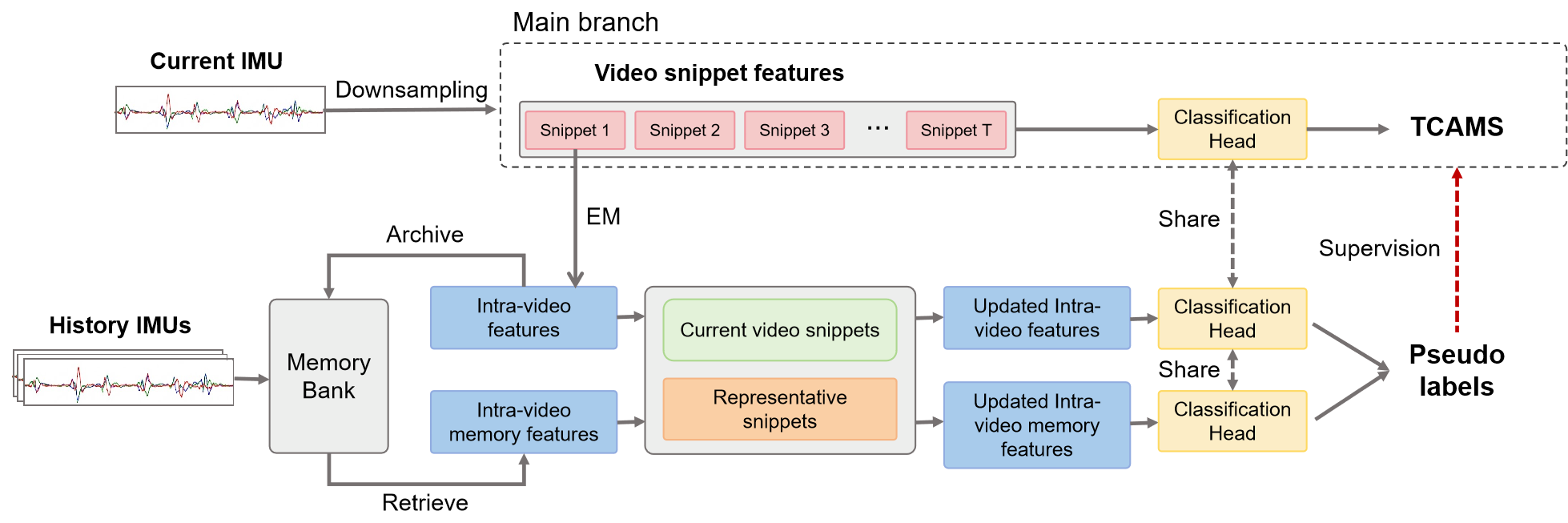}
    \caption{RSKP model architecture. A memory-augmented framework that utilizes history IMU data and a memory bank to generate pseudo labels, providing self-supervision for IMU-based temporal action recognition and classification.}
    \label{fig:RSKP_model}
\end{figure}

RSKP~\cite{huang2022weakly} refines coarse predictions by propagating high-confidence evidence to temporally related snippets/proposals\ref{fig:RSKP_model}.
Concretely, one constructs a temporal affinity matrix $\mathbf{A}\in\mathbb{R}^{P\times P}$ among $P$ proposals (or aggregated snippet groups) and performs iterative score refinement.
Let $\mathbf{S}^{(0)}$ be the initial proposal score matrix; the refinement runs for $T$ steps:
\begin{equation}
    \mathbf{S}^{(t+1)} = (1-\alpha)\mathbf{S}^{(0)} + \alpha\,\tilde{\mathbf{A}}\,\mathbf{S}^{(t)},
\end{equation}
where $\tilde{\mathbf{A}}$ is a normalized affinity matrix and $\alpha$ is a momentum hyperparameter.
This post-hoc propagation explicitly injects temporal smoothness/consistency, tends to fill ``holes'' inside an action, and reduces fragmented detections.

\textbf{Adaptations for Sequential IMU Data.}
The above strategies transfer naturally because an IMU stream admits the same snippet-based decomposition: we treat each short IMU window as a snippet instance and use a temporal encoder to produce $\Phi$.
We apply the CoLA-style contrastive loss directly on $\Phi$ \emph{before} MIL aggregation to encourage intra-action feature coherence under sensor noise and inter-subject variation.
Then, after obtaining proposal/snippet scores, we apply the RSKP-style temporal propagation as a light-weight refinement module operating purely in the time domain.

\section{Evaluation and Results}\label{sec:evaluation}

\subsection{Benchmarking Datasets}

\benchName is evaluated on seven datasets, including SBHAR~\cite{reyes2016transition}, Opportunity~\cite{roggen2010collecting},
WetLab~\cite{scholl2015wearables},
Hang-Time~\cite{hoelzemann2023hang},
RWHAR~\cite{sztyler2016body},
WEAR~\cite{bock2024wear}, and XRFV2~\cite{lan2025xrf}, enabling systematic benchmarking of weakly supervised IMU-TAL methods.
The first six datasets follow the benchmark selection adopted in the fully supervised IMU-TAL study by Bock et al.~\cite{bock2024temporal}, while XRFV2~\cite{lan2025xrf} is a more recent dataset released after Bock et al.~\cite{bock2024temporal} specifically to evaluate IMU-TAL.
Overall, this benchmark suite spans diverse domains, including sports-related activities (Hang-Time~\cite{hoelzemann2023hang} and WEAR~\cite{bock2024wear}), activities of daily living (Opportunity~\cite{roggen2010collecting}, SBHAR~\cite{reyes2016transition}, RWHAR~\cite{sztyler2016body}, and XRFV2~\cite{lan2025xrf}), and laboratory procedures (WetLab~\cite{scholl2015wearables}). Such diversity is important for assessing robustness, as the datasets vary substantially in terms of motion patterns, temporal structure, and execution variability across participants.

\begin{figure}[h]
    \centering
    \includegraphics[width=1\linewidth]{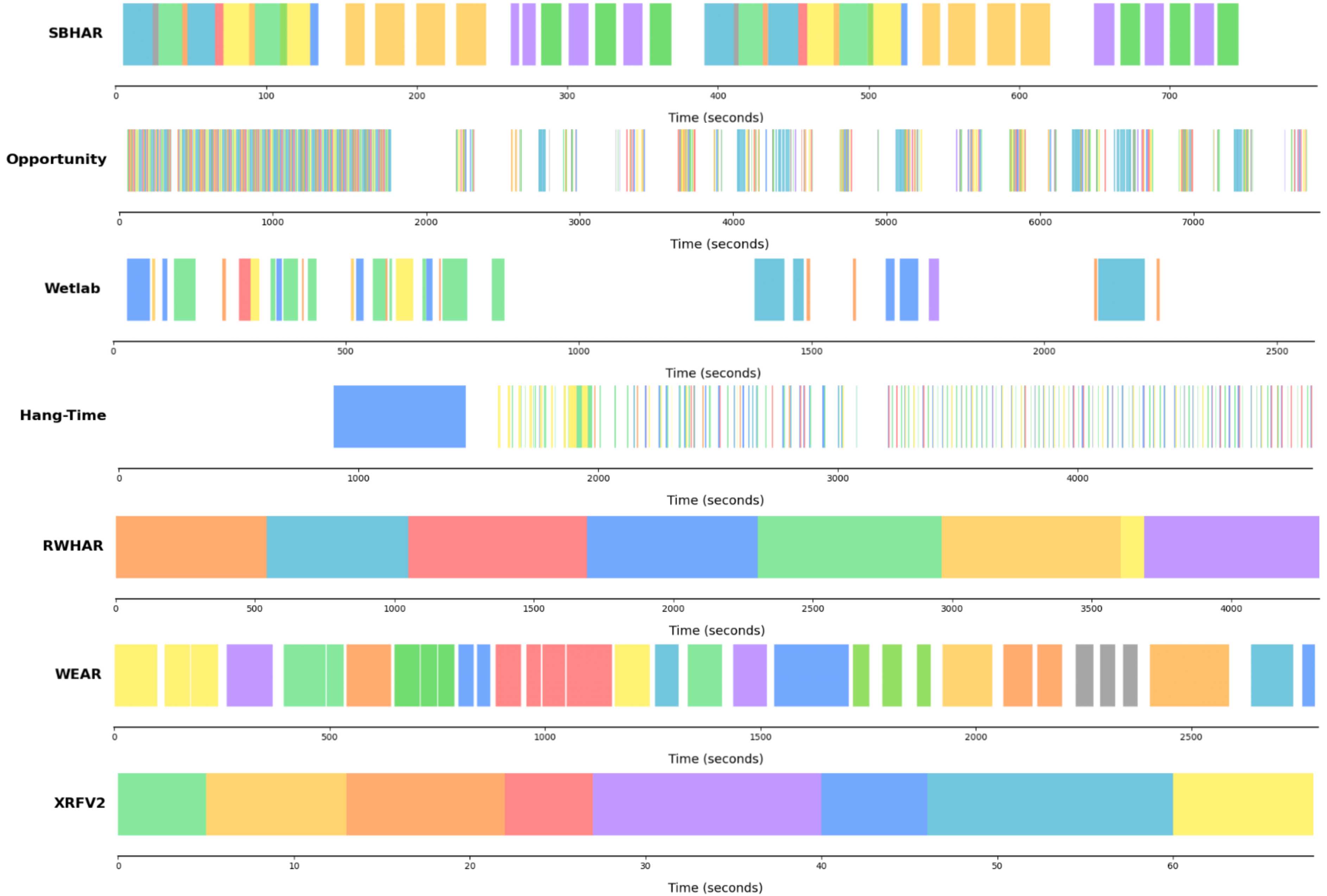}
    \caption{Dataset visualization(Opportunity~\cite{roggen2010collecting}, 
    SBHAR~\cite{reyes2016transition}, 
     WetLab~\cite{scholl2015wearables},
     Hang-Time~\cite{hoelzemann2023hang}, 
     RWHAR~\cite{sztyler2016body},
    WEAR~\cite{bock2024wear},
    and XRFV2~\cite{lan2025xrf}. ). Each subplot shows a continuous action sequence from a single subject over time, where the x-axis represents time and different colors denote different action classes.}
    \label{fig:dataset}
\end{figure}

The visualization of the datasets, shown in Figure~\ref{fig:dataset}, 
highlights the diverse temporal profiles across the benchmark. In the figure, each dataset is illustrated with one example action sequence, where different colors denote different action classes.
Overall, these datasets can be roughly grouped into three categories. The first category, including Opportunity and Hang-Time, is characterized by high temporal dynamics, featuring short individual action durations and frequent class transitions. The second category, including RWHAR and XRFV2, exhibits more stable temporal patterns; RWHAR, in particular, stands out with exceptionally long, continuous action segments, while XRFV2 presents relatively uniform action durations. The third category, comprising SBHAR, WetLab, and WEAR, strikes a balance, where action switches within a sequence appear more structured and evenly paced.

Table~\ref{tab:dataset_profile} summarizes the key dataset characteristics considered in our study. In particular, we report the number of subjects, the number of activity classes, and the number of sensor axes, which jointly reflect the scale of the benchmark, label granularity, and sensing dimensionality. 
We further report segment-duration statistics (in seconds) using Min/P5/Median/P95/Max, where Min/Median/Max denote the shortest/median/longest segments and P5/P95 are the 5th/95th percentiles (i.e., 5\%/95\% of segments have duration no greater than P5/P95). These statistics provide a compact view of temporal variability and help contextualize model behavior under short versus long activity instances. 
Notably, the wide gaps between the lower and upper percentiles in several datasets suggest heavy-tailed duration distributions, where most segments are short but a non-negligible fraction are substantially longer. This property can be particularly challenging for weakly supervised IMU-TAL, as it stresses both long-range temporal modeling and proposal generation/post-processing across heterogeneous time scales.


\begin{table}[t]
\centering
\setlength{\tabcolsep}{5pt}
\renewcommand{\arraystretch}{1.15}
\small
\begin{tabular}{lccc ccccc l}
\hline
\multirow{2}{*}{Dataset} &
\multirow{2}{*}{Subjects} &
\multirow{2}{*}{Classes} &
\multirow{2}{*}{Axes} &
\multicolumn{5}{c}{Segment duration (seconds)} &
\multirow{2}{*}{Scenario} \\
\cmidrule(lr){5-9}
& & & & Min & P5 & Median & P95 & Max & \\
\hline
SBHAR~\cite{reyes2016transition} & 30 & 12 & 3   & 1.46    & 2.74     & 12.76   & 26.57 & 40.62 & Locomotion \\
Opportunity~\cite{roggen2010collecting} & 4  & 17 & 113 & 0.23 & 1.20 & 2.60 & 6.7  & 17.07  & Daily-life actions \\
Wetlab~\cite{scholl2015wearables} & 22 & 8 & 3   & 0.28    & 1.48     & 11.23   & 64.54  & 214.38 & Laboratory \\
Hang-Time~\cite{hoelzemann2023hang} & 24 & 5 & 3   & 0.26 & 0.82    & 1.80  & 8.46  & 581.64  & Sports \\
RWHAR~\cite{sztyler2016body} & 15 & 8 & 21  & 84.14 & 98.26    & 621.16  & 691.61  & 1158.14  & Locomotion \\
WEAR~\cite{bock2024wear} & 18 & 18 & 12  & 1.36  & 9.11     & 43.12   & 104  & 183.10 & Sports \\
XRFV2~\cite{lan2025xrf} & 16  & 30 & 36 & 4.72 & 5.00 & 7.00 & 11.00  & 15.00  & Daily-life actions \\
\hline
\end{tabular}
\caption{Key dataset characteristics. We report the number of subjects, activity classes, sensor axes, and summary statistics of segment duration, together with the overall scenario. For segment duration, Min/Median/Max denote the shortest/median/longest segment; P5 and P95 are the 5th and 95th percentiles (i.e., 5\% and 95\% of segments have duration $\leq$ P5 and $\leq$ P95, respectively).}
\label{tab:dataset_profile}
\end{table}

\subsection{Benchmarking Methods}

As described in Section~\ref{sec:architecture-overview}, \benchName benchmarks seven representative weakly supervised localization methods adapted from three established modalities: weakly supervised sound event detection (WSSED):  DCASE~\cite{turpault2019sound}, CDur~\cite{dinkel2021towards}, weakly supervised object detection (WSOD): WSDDN~\cite{bilen2016weakly}, OICR~\cite{tang2017multiple}, PCL~\cite{tang2018pcl}, and weakly supervised temporal action localization in videos (WSVAL):  CoLA~\cite{zhang2021cola}, RSKP~\cite{huang2022weakly}.

These seven methods are widely recognized as canonical baselines in their respective communities, and together cover diverse weakly supervised learning principles (MIL pooling, proposal-based instance selection, pseudo-label refinement, contrastive learning, and knowledge propagation). This design allows \benchName to provide a systematic and reproducible benchmarking view on \emph{what transfers well}, \emph{what breaks}, and \emph{why} when porting weak supervision from audio, image, and video to IMU-based TAL.

\subsection{Implementation}\label{sec:implementation}

\emph{\textbf{(1) IMU Encoder Pretraining}}


Following common practice in WSOD and WSVAL, we first pretrain an IMU encoder to extract discriminative features that are later reused for proposal generation and weakly supervised localization.
To increase the amount of training data and to standardize the input length, we follow the protocol used in XRFV2~\cite{lan2025xrf} and segment each continuous sensor stream into fixed-length clips using a sliding-window scheme. Specifically, given a window length $W$, we extract overlapping clips with a 50\% stride (i.e., stride $=0.5W$), and treat each clip as a training sample under weak supervision. Unlike images and videos, where large-scale pretraining has produced general-purpose backbones~\cite{bilen2016weakly,tang2017multiple,tang2018pcl} that transfer well across downstream tasks, IMU signals exhibit strong subject- and dataset-specific characteristics (e.g., sensor placement, motion styles, and sampling frequency). Therefore, instead of relying on generic off-the-shelf pretraining, we initialize our IMU encoder with a module pretrained on the corresponding target dataset. This design better preserves dataset-specific invariances and reduces the risk of negative transfer across heterogeneous sensing conditions.

\emph{\textbf{(2) Evaluation Protocol.}} We adopt a leave-one-subject-out (LOSO) cross-validation protocol on Opportunity, Hang-Time, SBHAR, RWHAR, WetLab, and WEAR: we hold out one subject for testing and train on the remaining subjects. This process is repeated until each subject has served as the test split once, and we report the average performance.
For XRFV2, we evaluate under two complementary settings. (i) \textbf{Cross-subject (LOSO):} we randomly select four subjects and perform LOSO evaluation on them. (ii) \textbf{In-domain (no cross-subject):} following the official XRFV2 in-domain split, we additionally report results without cross-subject generalization. We evaluate the 7 above described weakly supervised methods and 3 fully supervised baselines (ActionFormer~\cite{zhang2022actionformer}, TemporalMaxer~\cite{tang2023temporalmaxer}, and TriDet~\cite{shi2023tridet}) on all these 7 datasets. Each method is run with three random seeds (2022, 2024, 2026).
Under this setup, the total number of training runs per dataset is computed as \#Seeds$\times$\#Subjects$\times$\#Models (e.g., for SBHAR: $3\times 30\times 10 = 900$). The full accounting is summarized in Table~\ref{tab:exp_accounting}. In total, \benchName comprises \textbf{3540} training runs, enabling systematic and reproducible benchmarking.

\begin{table}[h]
\centering
\setlength{\tabcolsep}{6pt}
\renewcommand{\arraystretch}{1.15}
\small
\begin{tabular}{lcccc}
\hline
Dataset & \#Seeds & \#Subjects & \#Models & Total runs \\
\hline
SBHAR & 3 & 30 & 7+3 & 900  \\
Opportunity & 3 & 4 & 7+3 & 120  \\
WetLab & 3 & 22 & 7+3 & 660  \\
Hang-Time & 3 & 24 & 7+3 & 720  \\
RWHAR & 3 & 15 & 7+3 & 450  \\
WEAR & 3 & 18 & 7+3 & 540 \\
XRFV2 & 3 & $4{+}1$ & 7+3 & 150  \\
\hline
\multicolumn{4}{l}{Total} & 3540  \\
\hline
\end{tabular}
\caption{Experiment accounting table. We evaluate 7 datasets with 10 models (7 weakly supervised + 3 fully supervised), using 3 random seeds (2022, 2024, 2026). The total number of runs is computed as \#Seeds$\times$\#Subjects$\times$\#Models.}
\label{tab:exp_accounting}
\end{table}

\emph{\textbf{(2) Inference.}}
To comprehensively evaluate model performance under varied input conditions, we design two inference modes: \texttt{full\_input} and \texttt{window\_input}.
The \texttt{full\_input} mode processes the entire continuous sensor stream as a single input, simulating an offline setting where the full sequence is available. Conversely, \texttt{window\_input} simulates an online/streaming setting by first dividing the test sequence into shorter windows. Each window is processed independently, and the resulting segment predictions are merged with temporal non-maximum suppression (NMS).
To keep \texttt{window\_input} meaningfully different from \texttt{full\_input} across datasets, we set the window length based on typical action durations. Specifically, for datasets dominated by shorter actions (Opportunity and Hang-Time), we use a 100-second window; for datasets with longer action instances (RWHAR, SBHAR, WEAR, and WetLab), we use a 1000-second window; and for XRFV2 we follow Lan et al.~\cite{lan2025xrf} and use a 30-second window.

By comparing \texttt{full\_input} and \texttt{window\_input}, we can assess model behavior under both complete-stream and segmented-stream conditions.

\emph{\textbf{(3) Hyperparameter Settings.}}
Our experiments are conducted in two stages. In the pretraining stage, we use the CosineAnnealingLR optimizer~\cite{loshchilov2017sgdr} with an initial learning rate of 1e-3 and a weight decay coefficient of 5e-5, training for a total of 60 epochs. In the subsequent training stage, we use the Adam optimizer~\cite{kingma2015adam} with a learning rate of 1e-4 and a weight decay coefficient of 1e-5. The learning rate is decayed by a factor of 0.9 every 10 epochs, and the model is trained for 80 epochs in total. 
Given the unique characteristics of IMU data compared to audio, images, and video, we have made some adjustments to the model's architecture to better adapt it to IMU signals. Additionally, since the Selective Search algorithm~\cite{van2011segmentation} commonly used in weakly supervised learning for image data is not suitable for IMU data, we designed an alternative algorithm to achieve similar functionality, as described in Algorithm \ref{alg:proposal_boxes}. 

\emph{\textbf{(4) Post-processing.}}
During the inference phase, the model generates initial action outputs for each input sequence. These outputs are then refined using temporal NMS to merge overlapping predictions. The final output consists of a tuple of action outputs, $\{c$, $c_{score}$, $[t_{\text{start}}, t_{\text{end}}]\}$, where $c$ represents the predicted action class, $c_{score}$ is the associated confidence score, and $[t_{\text{start}}, t_{\text{end}}]$ denotes the corresponding start and end boundaries.

\begin{table}[h]
\centering
\small
\setlength{\tabcolsep}{6pt}
\renewcommand{\arraystretch}{1.15}
\begin{tabular}{lll}
\hline
\textbf{Setting} & \textbf{Datasets} & \textbf{Value} \\
\hline
\texttt{window\_input} window length & Opportunity, Hang-Time & 100 seconds \\
\texttt{window\_input} window length & RWHAR, SBHAR, WEAR, WetLab & 1000 seconds \\
\texttt{window\_input} window length & XRFV2 & 30 seconds \\
Random seeds & All & 2022, 2024, 2026 \\
Number of proposals per sequence  & All & 3000 \\
Segment-level mAP thresholds & All & $\tau\in\{0.3,0.4,0.5,0.6,0.7\}$ \\
\hline
\end{tabular}
\caption{Fixed settings used throughout our benchmark for reproducibility.}
\label{tab:repro_settings}
\end{table}

\emph{\textbf{(5) Reproducibility Checklist.}}
To reduce ambiguity and make our benchmark easier to reproduce and extend, we summarize below the key implementation choices that are fixed across all experiments. Our code will be made publicly available. 

\begin{itemize}[leftmargin=*]
    \item \textbf{Training clip generation (all methods).} Each continuous stream is segmented into fixed-length clips using a sliding window of length $W$ with a 50\% stride (stride $=0.5W$).

    \item \textbf{Evaluation splits (all methods).} We use LOSO cross-validation on Opportunity, Hang-Time, SBHAR, RWHAR, WetLab, and WEAR. For XRFV2, we report both (i) cross-subject (LOSO on 4 selected subjects) and (ii) in-domain (official split).

    \item \textbf{Inference modes (all methods).} We evaluate both \texttt{full\_input} (entire stream) and \texttt{window\_input} (stream partitioned into windows and merged).

    \item \textbf{Window length for \texttt{window\_input}.} We fix the window length by dataset characteristics as summarized in Table~\ref{tab:repro_settings}.

    \item \textbf{Proposal generation for WSOD-derived methods.} For proposal-based methods (WSDDN/OICR/PCL), we generate 3000 temporal proposals per sequence using the multi-scale sampling procedure in Algorithm~\ref{alg:proposal_boxes}. This procedure is parameterized by the duration settings $(\text{min\_sec}, \text{max\_sec}, \text{sec\_resolution})$ and the mix of structured versus random sampling controlled by \text{fixed\_keep\_ratio}.

    \item \textbf{Optimization (all methods).} We use a two-stage optimization schedule: (i) encoder pretraining for 60 epochs with CosineAnnealingLR (initial LR $=10^{-3}$, weight decay $=5\times 10^{-5}$), then (ii) weakly supervised training for 80 epochs with Adam (LR $=10^{-4}$, weight decay $=10^{-5}$), decaying the LR by $\times 0.9$ every 10 epochs.

    \item \textbf{Post-processing (all methods).} We apply temporal NMS to merge overlapping predictions, and output segment tuples $(c, c_{score}, [t_{\text{start}}, t_{\text{end}}])$. For visualization, we additionally apply a deterministic merge rule (keep the highest-scoring label for overlaps; merge touching/overlapping segments of the same label).
\end{itemize}

\subsection{Metrics}\label{sec:metrics}


We evaluate WS-IMU-TAL from two complementary views: \emph{(i) frame-level labeling} measured by per-frame (IMU sampling point) classification metrics (Precision, Recall, and F1) and misalignment ratios (UODIFM), and \emph{(ii) segment-level detection} measured by mAP under temporal Intersection-over-Union (tIoU) thresholds.

Following the evaluation procedure in~\cite{bock2024temporal}, the frame-level metrics are computed by first converting
both predicted segments and ground-truth segments (GT) into a discrete label sequence over time, and then evaluating the
resulting sequences. Here, the discrete label sequence is obtained by rasterizing each segment tuple $(c, [t_{\text{start}}, t_{\text{end}}])$ into per-sample class labels along the timeline.

\emph{\textbf{(1) Per-frame Precision, Recall, and F1 (frame-level).}}
We evaluate the model as a \emph{continuous labeling} problem.
Given the sampling rate, we convert both ground-truth annotations and predicted segments into two discrete label sequences
$\{y_t\}_{t=1}^{T}$ and $\{\hat{y}_t\}_{t=1}^{T}$, where $y_t,\hat{y}_t\in\{0,1,\dots,C\}$
denote the class index at time step $t$.
Following~\cite{bock2024temporal}, a time step is assigned to a class if it falls inside
a predicted/GT segment of that class; otherwise it is assigned to \texttt{null}.
We then compute per-class frame-level Precision, Recall, and F1 score:
\begin{equation}\label{eq:precision}
\text{Precision}_c = \frac{TP_c}{TP_c + FP_c},
\end{equation}
\begin{equation}\label{eq:recall}
\text{Recall}_c = \frac{TP_c}{TP_c + FN_c},
\end{equation}
\begin{equation}\label{eq:f1-score}
\text{F1}_c = 2 \cdot \frac{\text{Precision}_c \cdot \text{Recall}_c}{\text{Precision}_c + \text{Recall}_c},
\end{equation}
where $TP_c,FP_c,FN_c$ are counted over time steps for class $c$.
We report the average across classes (excluding \texttt{null} when present), yielding dataset-level frame-wise Precision, Recall, and F1.

\emph{\textbf{(2) Misalignment Ratios (UODIFM, frame-level).}}
To diagnose precisely \emph{how} predictions are misaligned in time, beyond what a single F1-score can reveal, we adopt the detailed error characterization from Ward et al.~\cite{ward2006evaluating}. Following the implementation in~\cite{bock2024temporal}, this method operates on the rasterized (frame-level) label sequences and measures six distinct types of alignment errors. These errors are grouped into two categories:

\textbf{Ground-Truth-Side Errors (What the model missed):}
1) Deletion (DR): The model completely failed to detect a ground-truth action instance. The error is the total duration of all such missed instances.
2) Underfill (UR): The model detected an action but its prediction did not cover the full duration, missing either the beginning or the end. The error is the total duration of these missed boundary portions.
3) Fragmentation (FR): The model detected a single continuous action as two or more separate segments, leaving gaps in between. The error is the total duration of these internal gaps.

\textbf{Prediction-Side Errors (What the model hallucinated):}
4) Insertion (IR): The model predicted an action where there was only background. The error is the total duration of these false predictions.
5) Overfill (OR): The model's prediction extended beyond the boundaries of a true action into the background. The error is the total duration of these ``spilled over" portions.
6) Merge (MR): The model's prediction was so long that it incorrectly merged two or more distinct ground-truth actions into a single one. The error is the duration of the incorrect prediction that bridges the true actions.

For each class, these accumulated error durations are normalized to produce a ratio. Following the established implementation, we then average these ratios across all non-background classes and present the final values as percentages ($\times 100$).

\emph{\textbf{(3) tIoU, AP and mAP (segment-level).}}
We use tIoU, widely-applied in~\cite{zhang2022actionformer,tang2023temporalmaxer,shi2023tridet}, 
to quantify the overlap between a predicted segment $p=[s_{\text{pred}},e_{\text{pred}}]$ and
a ground-truth segment $g=[s_{\text{gt}},e_{\text{gt}}]$.
For a given threshold $\tau$, a prediction is considered correct if it has the same class as a ground-truth segment
and achieves $tIoU\ge\tau$.
Average Precision (AP) at threshold $\tau$, denoted as $AP@\tau$, is computed from the precision--recall curve obtained by
ranking predictions by confidence scores. We report mean AP (mAP) averaged over
$\tau\in\{0.3,0.4,0.5,0.6,0.7\}$.

\begin{table*}[!htbp]
  \centering
  \scriptsize

  \begin{tabular}{@{}llccc|cccccc|c@{}}
    & Model & P ($\uparrow$) & R ($\uparrow$) & F1 ($\uparrow$) & UR ($\downarrow$) & OR ($\downarrow$) & DR ($\downarrow$) & IR ($\downarrow$) & FR ($\downarrow$) & MR ($\downarrow$) & mAP ($\uparrow$) \\ 
    \midrule

    \rowcolor{lightgray} & ActionFormer & 87.93 & 84.17 & 84.83 & 0.39 & 5.39 & 0.14 & 5.25 & 0.00 & 0.21 & 94.46 \\
    \rowcolor{lightgray} & TemporalMaxer & 87.87 & 83.56 & 84.50 & 0.39 & 5.97 & 0.18 & 4.57 & 0.00 & 0.24 & 94.48 \\
    \rowcolor{lightgray} & TriDet & \textbf{88.63} & \textbf{86.06} & \textbf{86.23} & 0.38 & 5.39 & 0.13 & 4.29 & 0.01 & 0.28 & \textbf{94.74} \\
    \rowcolor{lightgreen} & DCASE & 6.01 & 6.50 & 5.24 & 0.60 & 5.29 & 5.62 & 39.65 & 0.03 & 0.42 & 1.06 \\
    \rowcolor{lightgreen} & CDur & 5.48 & 6.28 & 5.44 & 0.38 & 5.77 & 5.87 & 57.23 & 0.02 & 0.35 & 0.92 \\
    \rowcolor{lightorange} & WSDDN & 13.55 & 8.62 & 5.24 & 0.55 & 6.57 & 5.36 & 85.67 & 0.06 & 0.21 & 2.45 \\
    \rowcolor{lightorange} & OICR & 3.68 & 5.99 & 3.16 & 0.42 & 7.68 & 5.75 & 80.69 & 0.02 & 0.43 & 0.98 \\
    \rowcolor{lightorange} & PCL & 4.14 & 5.57 & 3.59 & 0.37 & 7.65 & 5.80 & 80.45& 0.02 & 0.28 & 1.24 \\
    \rowcolor{lightred} & RSKP & \textbf{35.66} & \textbf{36.29} & \textbf{28.76} & 0.33 & 12.74 & 1.59 & 9.31 & 0.01 & 3.77 & \textbf{37.16} \\

    \rowcolor{lightred}    \multirow{-10}{*}{\rotatebox[origin=c]{90}{\textbf{SBHAR}}}    & CoLA & 15.92 & 16.92 & 14.38 & 1.14 & 3.24 & 3.27 & 64.34 & 0.51 & 0.03 & 12.57 \\
   \hline

    \rowcolor{lightgray}
    & ActionFormer & \textbf{54.36} & \textbf{58.67} & 51.67 & 0.19 & 13.14 & 0.42 & 34.26 & 0.00 & 0.47 & \textbf{50.82} \\
  \rowcolor{lightgray}
    & TemporalMaxer & 44.75 & 55.70 & 44.80 & 0.21 & 14.17 & 0.39 & 43.48 & 0.01 & 0.49 & 46.09 \\
   
   \rowcolor{lightgray}
    & TriDet & 49.14 & 57.83 & 48.93 & 0.23 & 13.01 & 0.38 & 39.84 & 0.00 & 0.56 & 49.80 \\ 
    \rowcolor{lightgreen}
    & DCASE & \textbf{48.10} & 22.82 & \textbf{24.33} & 0.33 & 9.66 & 0.90 & 32.23 & 0.04 & 0.12 & 9.14 \\
    \rowcolor{lightgreen}
    & CDur & 41.96 & 23.35 & 24.07 & 0.27 & 11.75 & 0.98 & 37.26 & 0.03 & 0.34 & \textbf{10.71} \\
    \rowcolor{lightorange}
    & WSDDN & 14.76 & 21.74 & 15.19 & 0.13 & 25.72 & 1.13 & 62.53 & 0.00 & 0.45 & 3.05 \\
    \rowcolor{lightorange} 
    & OICR & 18.11 & 19.96 & 16.19 & 0.10 & 31.73 & 1.21 & 50.77 & 0.01 & 0.45 & 2.50\\
    \rowcolor{lightorange} 
    & PCL & 18.24 & 20.66 & 16.83 & 0.10 & 35.01 & 1.22 & 47.89 & 0.00 & 0.42 & 2.55 \\
    \rowcolor{lightred}
    & RSKP & 23.70 & \textbf{35.84} & 19.15 & 0.13 & 29.34 & 0.73 & 46.93 & 0.003 & 0.83 & 7.53 \\
    \rowcolor{lightred}    \multirow{-10}{*}{\rotatebox[origin=c]{90}{\textbf{Opportunity}}}   & CoLA & 10.55 & 21.42 & 10.18 & 0.36 & 3.56 & 0.78 & 89.48 & 0.12 & 0.03 & 4.09 \\ \hline

    \rowcolor{lightgray} & ActionFormer & 37.6 & 50.29 & 38.17 & 0.56 & 9.94 & 0.82 & 51.99 & 0.08 & 0.75 & 36.95 \\
    \rowcolor{lightgray} & TemporalMaxer & 36.20 & 49.70 & 36.55 & 0.59 & 9.62 & 0.86 & 55.31 & 0.11 & 0.61 & 36.16 \\
    \rowcolor{lightgray} & TriDet & \textbf{40.10} & \textbf{50.38} & \textbf{40.04} & 0.56 & 8.86 & 0.82 & 49.50 & 0.09 & 0.79 & \textbf{37.14} \\ 
    \rowcolor{lightgreen} & DCASE & 27.41 & 26.22 & 22.62 & 0.71 & 1.09 & 1.46 & 44.24 & 0.36 & 0.01 & 9.68 \\
    \rowcolor{lightgreen} & CDur & \textbf{45.35} & 29.02 & \textbf{30.82} & 0.85 & 1.31 & 1.00 & 30.78 & 0.46 & 0.06 & 12.60 \\
    \rowcolor{lightorange} & WSDDN & 31.51 & 26.59 & 15.79 & 0.67 & 10.23 & 1.76 & 63.70 & 0.11 & 0.66 & 6.48 \\
    \rowcolor{lightorange} & OICR & 31.97 & 41.63 & 22.41 & 0.59 & 19.82 & 1.10 & 49.60 & 0.08 & 1.42 & 14.17 \\
    \rowcolor{lightorange} & PCL & 31.81 & \textbf{43.78} & 23.48 & 0.54 & 22.30 & 1.05 & 49.45 & 0.04 & 1.75 & \textbf{16.27} \\
    \rowcolor{lightred} & RSKP & 29.80 & 42.54 & 29.25 & 0.62 & 2.76 & 0.34 & 68.33 & 0.73 & 0.005 & 14.75 \\
    \rowcolor{lightred}     \multirow{-10}{*}{\rotatebox[origin=c]{90}{\textbf{Wetlab}}}& CoLA & 16.96 & 24.40 & 9.89 & 0.80 & 4.51 & 1.56 & 88.12 & 0.22 & 0.26 & 10.32 \\
  \hline


    \rowcolor{lightgray} & ActionFormer & 49.09 & \textbf{57.38} & \textbf{51.16} & 0.62 & 11.70 & 0.49 & 47.76 & 0.50 & 0.59 & 29.30 \\
    \rowcolor{lightgray} & TemporalMaxer & 45.18 & 54.68 & 47.40 & 0.70 & 11.04 & 0.45 & 52.43 & 1.12 & 0.65 & 27.98 \\
    \rowcolor{lightgray} & TriDet & \textbf{49.61} & 55.13 & 50.67 & 0.71 & 9.83 & 0.53 & 48.56 & 0.70 & 0.61 & \textbf{29.35} \\ 
    \rowcolor{lightgreen} & DCASE & 59.92 & \textbf{55.22} & 53.30 & 0.62 & 9.38 & 0.51 & 37.27 & 0.48 & 0.16 & \textbf{27.14} \\
    \rowcolor{lightgreen} & CDur & \textbf{71.00} & 52.68 & \textbf{56.78} & 0.65 & 8.33 & 0.50 & 23.05 & 0.52 & 0.12 & 24.56 \\
    \rowcolor{lightorange} & WSDDN & 37.56 & 45.82 & 35.09 & 0.48 & 21.36 & 1.28 & 49.89 & 1.29 & 1.20 & 4.97 \\
    \rowcolor{lightorange} & OICR & 27.38 & 41.55 & 25.61 & 0.57 & 22.39 & 1.07 & 60.03 & 1.51 & 1.29 & 3.28 \\
    \rowcolor{lightorange} & PCL & 28.56 & 45.41 & 27.09 & 0.48 & 24.63 & 1.00 & 56.65 & 1.11 & 1.65 & 3.11 \\
    \rowcolor{lightred} & RSKP & 32.86 & 46.73 & 31.04 & 0.44 & 9.56 & 0.45 & 69.56 & 0.34 & 0.22 & 15.94 \\
    \rowcolor{lightred}    \multirow{-10}{*}{\rotatebox[origin=c]{90}{\textbf{Hang-Time}}} & CoLA & 16.93 & 32.23 & 13.90 & 0.92 & 2.76 & 0.49 & 88.12 & 2.94 & 0.02 & 7.87 \\
   \hline

    \rowcolor{lightgray} & ActionFormer & 59.31 & 61.85 & 56.53 & 2.40 & 10.35 & 1.64 & 11.08 & 0.12 & 0.00 & 63.81 \\
    \rowcolor{lightgray} & TemporalMaxer & 54.91 & 60.35 & 54.25 & 2.27 & 12.59 & 1.65 & 14.17 & 0.44 & 0.00 & 48.15 \\
    \rowcolor{lightgray} & TriDet & \textbf{62.29} & \textbf{65.07} & \textbf{60.31} & 1.49 & 6.94 & 1.77 & 9.50 & 0.18 & 0.00 & \textbf{70.85} \\ 
    \rowcolor{lightgreen} & DCASE & \textbf{64.37} & \textbf{64.44} & \textbf{61.76} & 0.94 & 2.59 & 1.82 & 13.60 & 0.71 & 0.00 & \textbf{65.07} \\
    \rowcolor{lightgreen} & CDur & 56.20 & 41.67 & 43.17 & 1.36 & 1.08 & 1.88 & 18.91 & 2.92 & 0.00 & 24.69 \\
    \rowcolor{lightorange} & WSDDN & 46.51 & 47.96 & 43.13 & 3.65 & 21.92 & 2.29 & 16.43 & 0.29 & 0.18 & 36.22 \\
    \rowcolor{lightorange} & OICR & 55.29 & 58.53 & 53.30 & 2.55 & 12.90 & 1.32 & 14.57 & 0.71 & 0.32 & 31.24 \\
    \rowcolor{lightorange} & PCL & 55.77 & 59.66 & 54.09 & 2.15 & 11.73 & 1.48 & 15.39 & 0.74 & 0.09 & 28.73 \\
    \rowcolor{lightred} & RSKP & 63.73 & 64.34 & 59.80 & 1.09 & 1.49 & 1.41 & 17.78 & 1.14 & 0.00 & 53.93 \\
    \rowcolor{lightred}\multirow{-10}{*}{\rotatebox[origin=c]{90}{\textbf{RWHAR}}} & CoLA & 14.81 & 14.16 & 10.13 & 4.41 & 3.35 & 6.65 & 78.33 & 1.17 & 0.00 & 18.88 \\
   \hline

    \rowcolor{lightgray} & ActionFormer & 72.02 & \textbf{76.93} & 72.56 & 0.21 & 6.39 & 0.64 & 6.90 & 0.01 & 2.07 & 74.45 \\
    \rowcolor{lightgray} & TemporalMaxer & 68.96 & 72.88 & 69.23 & 0.23 & 6.76 & 0.81 & 5.95 & 0.01 & 1.62 & 70.13 \\
    \rowcolor{lightgray} & TriDet & \textbf{74.91} & 75.67 & \textbf{73.14} & 0.29 & 4.63 & 0.63 & 6.51 & 0.02 & 1.47 & \textbf{75.37} \\ 
    \rowcolor{lightgreen} & DCASE & 68.31 & 62.06 & 59.89 & 0.28 & 2.72 & 0.58 & 19.34 & 0.54 & 0.38 & 30.78 \\
    \rowcolor{lightgreen} & CDur & \textbf{75.38} & 46.62 & 51.34 & 0.53 & 1.61 & 0.62 & 10.07 & 0.84 & 0.16 & 15.72 \\
    \rowcolor{lightorange} & WSDDN & 57.62 & 76.17 & 58.66 & 0.26 & 18.15 & 0.46 & 17.83 & 0.05 & 5.20 & 28.02 \\
    \rowcolor{lightorange} & OICR & 47.48 & 64.96 & 48.22 & 0.51 & 20.93 & 0.57 & 21.58 & 0.07 & 4.34 & 26.03 \\
    \rowcolor{lightorange} & PCL & 47.43 & 65.81 & 49.06 & 0.57 & 24.91 & 0.51 & 18.23 & 0.06 & 0.57 & 31.66 \\
    \rowcolor{lightred} & RSKP & 70.80 & \textbf{80.57} & \textbf{71.84} & 0.15 & 5.60 & 0.42 & 20.44 & 0.11 & 0.88 & \textbf{63.29} \\
    \rowcolor{lightred}  \multirow{-10}{*}{\rotatebox[origin=c]{90}{\textbf{WEAR}}} & CoLA & 17.28 & 29.44 & 16.23 & 0.54 & 2.63 & 1.81 & 81.25 & 0.14 & 0.30 & 24.38 \\
   \hline

    \rowcolor{lightgray} & ActionFormer & 56.46 & 51.51 & 51.05 & 0.70 & 9.23 & 0.46 & 15.46 & 0.05 & 0.00 & 80.33 \\
    \rowcolor{lightgray} & TemporalMaxer & \textbf{64.62} & \textbf{55.36} & \textbf{53.79} & 0.68 & 5.31 & 0.47 & 15.17 & 0.07 & 0.00 & \textbf{84.31} \\
    \rowcolor{lightgray} & TriDet & 58.54 & 50.55 & 50.28 & 0.58 & 5.45 & 0.08 & 23.62 & 0.05 & 0.00 & 76.04 \\ 
    \rowcolor{lightgreen} & DCASE & 20.28 & 20.34 & 17.27 & 0.64 & 27.68 & 1.94 & 11.20 & 0.00 & 0.17 & 17.02 \\
    \rowcolor{lightgreen} & CDur & 18.21 & 12.60 & 11.55 & 0.45 & 14.04 & 2.02 & 16.30 & 0.03 & 0.84 & 7.89 \\
    \rowcolor{lightorange} & WSDDN & 5.63 & 4.26 & 4.45 & 0.02 & 4.07 & 3.20 & 0.68 & 0.00 & 0.00 & 3.85 \\
    \rowcolor{lightorange} & OICR & 20.20 & 17.60 & 16.81 & 0.10 & 12.46 & 2.56 & 4.45 & 0.00 & 0.00 & 17.42 \\
    \rowcolor{lightorange} & PCL & 16.66 & 15.90 & 15.04 & 0.11 & 14.51 & 2.58 & 7.09 & 0.00 & 0.07 & 13.80 \\ 
    \rowcolor{lightred} & RSKP & 23.36 & \textbf{24.69} & 20.61 & 0.26 & 17.38 & 2.02 & 10.31 & 0.01 & 1.51 & \textbf{26.81} \\
    \rowcolor{lightred}     \multirow{-10}{*}{\rotatebox[origin=c]{90}{\textbf{XRFV2}}}& CoLA & \textbf{26.53} & 21.31 & \textbf{21.25} & 0.39 & 5.14 & 1.68 & 14.94 & 0.10 & 0.01 & 19.63 \\
    \hline
  \end{tabular}
    \caption{\textit{Action localization based on supervised learning and weakly supervised learning:} Average LOSO cross-validation results obtained on seven inertial HAR benchmark datasets \cite{roggen2010collecting, reyes2016transition, scholl2015wearables, sztyler2016body, bock2024wear, hoelzemann2023hang, lan2025xrf} for seven weakly supervised learning \cite{bilen2016weakly,tang2017multiple,tang2018pcl} and three supervised learning TAL architectures \cite{zhang2022actionformer, tang2023temporalmaxer, shi2023tridet}. The table provides per-sample classification metrics, i.e. Precision (P), Recall (R), F1-Score (F1), misalignment ratios \cite{ward2006evaluating} and average mAP applied at different tIoU thresholds (0.3,0.4,0.5,0.6,0.7). Best results per dataset are in \textbf{bold}.}
  \label{tab:result_supervised_weaklySupervised}
\end{table*}

\subsection{Results} \label{sec:results}

\paragraph{\textbf{Answering RQ2 (Effectiveness).}}
RQ2 asks how effective the transferred weakly supervised methods are for recognition and temporal localization on IMU data.
We answer RQ2 by benchmarking seven representative weakly supervised approaches (from audio, image, and video) on seven IMU-TAL datasets, and comparing them against fully supervised upper bounds under the same evaluation protocol.

\emph{\textbf{(1) Supervised vs. Weakly Supervised Performance on IMU Data.}}\label{sec:supervised_vs_weakly}
The experimental results in Table~\ref{tab:result_supervised_weaklySupervised} first establish the viability of weakly supervised learning for IMU Temporal Action Localization. The three rows highlighted in light gray report fully supervised baselines and serve as an approximate upper bound.
The remaining rows are grouped by their source modality: light green for WSSED (audio), light orange for WSOD (image), and light red for WSVTAL (video), enabling a clear comparison of how weak supervision transfers to the IMU domain.

Without requiring any frame-level annotations, weakly supervised models achieve meaningful localization performance, particularly on RWHAR and WEAR. On these two datasets, DCASE and RSKP reach mAP scores of 65.07\% and 63.29\%, respectively, substantially narrowing the gap to the best fully supervised methods. This strong performance is consistent with the dataset characteristics in Table~\ref{tab:dataset_profile}: both RWHAR and WEAR contain long action instances (median duration $>40$~s) and have relatively high-dimensional sensor inputs (21 and 12 axes, respectively), providing richer and more discriminative signals for learning under weak supervision.

This success, however, is not universal and depends strongly on the intrinsic properties of each dataset. All weakly supervised methods exhibit a pronounced performance drop on Opportunity and SBHAR.
For Opportunity, the main challenge is the prevalence of very short Activities of Daily Living (median duration $=2.6$~s), such as ``open/close door'', where stable and discriminative IMU patterns are difficult to capture within a limited temporal context.
SBHAR, in contrast, mainly consists of brief transitional actions (e.g., ``sit-to-stand'') and is recorded with low-dimensional (3-axis) sensor inputs, further constraining the available supervision signal.
Overall, these results indicate that weakly supervised methods are highly sensitive to both action duration and input dimensionality, whereas fully supervised approaches are substantially more robust in signal-sparse and data-limited regimes.

This performance bottleneck further highlights a clear modality-dependent transfer gap. Methods transferred from intrinsically temporal domains like video (e.g., RSKP) and audio (e.g., DCASE) tend to be stronger and more stable, likely because their core designs assume continuous, sequential streams that are structurally closer to IMU signals. In contrast, image-derived WSOD methods (WSDDN, OICR, and PCL) consistently underperform across datasets. A key limitation is their proposal generation mechanism, which we adapt to the temporal axis by mimicking region proposal strategies from computer vision. This procedure relies on stochastic, content-agnostic sampling and thus mismatches the strong temporal regularities of IMU activities (e.g., periodic patterns and structured transitions). As a result, the generated proposals are often temporally imprecise and fail to align with true action boundaries, which in turn propagates noise to subsequent instance selection, classification, and refinement stages, ultimately capping performance.

\begin{figure}[h]
    \centering
    \includegraphics[width=0.8\linewidth]{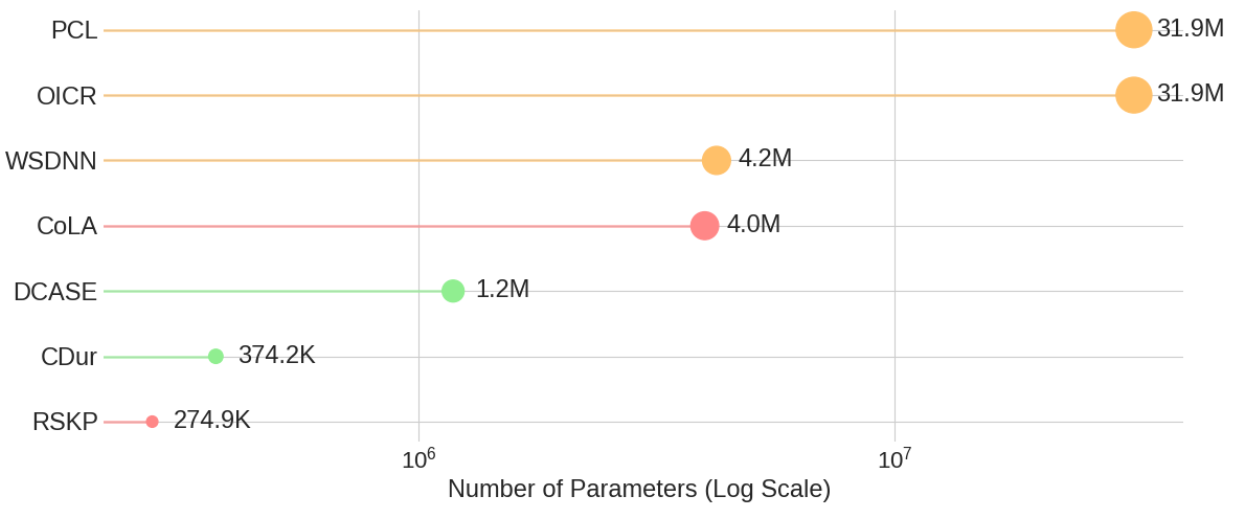}
    \caption{Comparison of the number of trainable parameters for each weakly supervised model.}
    \label{fig:parameter}
\end{figure}

\emph{\textbf{(2) Model Complexity}}\label{sec:complexity}
Figure~\ref{fig:parameter} illustrates the number of trainable parameters for each evaluated weakly supervised model, revealing significant disparities in their architectural complexity. The models can be broadly categorized into three groups: RSKP (274,944) and CDur (374,163) are exceptionally lightweight; DCASE, CoLA, and WSDDN form a middleweight tier; while OICR and PCL (both at 31,907,948) are heavyweight models with a massive number of parameters.

A key finding is the absence of a direct positive correlation between the number of parameters and model performance. For instance, the RSKP model, which has the fewest parameters, is a top performer across many of our experiments. This provides strong evidence that for the WS-IMU-TAL task, an efficient and well-suited architecture design is far more critical than simply scaling up the parameter count.

Furthermore, a notable observation from the figure is that the OICR and PCL models possess an identical number of parameters. This is by design, as both models share the exact same underlying neural network architecture. The distinction between them lies not in their structure, but entirely in their training-phase methodologies; specifically, they employ different pseudo-label generation algorithms and loss calculation strategies. This fact underscores that within the weakly supervised learning paradigm, the choice of training strategy and loss function can be as impactful as the network architecture itself.

\emph{\textbf{(3) Ablation Study: Window vs. Full-Sequence Input.}}
To investigate the impact of temporal context on weak supervision, we compare two input strategies: window-level  (fixed-length segments) and full-sequence (the entire recording). The results, in Table ~\ref{tab:input_result}, reveal two distinct patterns of behavior rooted in the models' underlying architectures, while some benefit from global context, others are hindered by it.

For detection-based models such as DCASE and CDur, full-sequence input generally yields better mAP. This is particularly evident in the WetLab and RWHAR datasets, where global context helps clarify similar motion patterns and refine action boundaries. Conversely, MIL-based frameworks like WSDDN, OICR, and PCL result in lower performance when processing full sequences. This stems from the MIL dilution effect, where the increase in sequence length erodes the signal-to-noise ratio of sparse IMU actions, resulting in degraded class activation maps.

Observations reveal that although adapted from video domains, CoLA and RSKP exhibit diverging behaviors. CoLA often favors window-level input because localized segments provide a cleaner search space for its contrastive learning mechanism, whereas full-sequence noise hinders effective negative mining. In contrast, RSKP demonstrates the highest robustness, leveraging its semantic prompting to maintain high-precision localization ($tIoU \ge 0.5$) across full-sequence benchmarks like SBHAR and WEAR. This indicates that while window-level input simplifies local feature extraction, global temporal reasoning remains essential for complex, real-world IMU streams.

\begin{table*}[h!]
  \centering
  \footnotesize

  \begin{tabular}{@{}llcccccc@{}}
    \hline
    Dataset & Model & tIoU-0.3 & tIoU-0.4 & tIoU-0.5 & tIoU-0.6 & tIoU-0.7 & mAP \\
    \hline

    \rowcolor{lightgreen} & DCASE & 1.66|2.26 & 1.33|1.74 & 0.73|0.79 & 0.44|0.39 & 0.19|0.11 & 0.87|1.06 \\
    \rowcolor{lightgreen} & CDur & 1.33|2.08 & 1.01|1.59 & 0.54|0.62 & 0.35|0.22 & 0.22|0.06 & 0.69|0.92 \\
    \rowcolor{lightorange} & WSDDN & 6.36|4.86 & 4.49|3.43 & 2.91|2.05 & 1.73|1.27 & 0.77|0.64 & 3.25|2.45\\
    \rowcolor{lightorange} & OICR & 1.98|2.00 & 1.32|1.35 & 0.84|0.88 & 0.39|0.46 & 0.20|0.23 & 0.95|0.98 \\
    \rowcolor{lightorange} & PCL & 2.12| 2.37 & 1.42|1.71 & 0.95|1.10 & 0.62|0.65 & 0.27|0.39 & 1.07|1.24 \\
    \rowcolor{lightred} & RSKP & \textbf{59.77}|\textbf{47.75} & \textbf{48.43}|\textbf{43.25} & \textbf{33.39}|\textbf{38.01} & \textbf{22.26}|\textbf{32.27} & \textbf{13.17}|\textbf{24.50} & \textbf{35.40}|\textbf{37.16} \\
    \rowcolor{lightred} \multirow{-7}{*}{\rotatebox[origin=c]{90}{SBHAR}}
    & CoLA & 18.30|17.37 & 15.74|15.98 & 12.59|13.47 & 8.82|10.29 & 5.14|5.76 & 12.12|12.57 \\
    \hhline{--------}

    \rowcolor{lightgreen} & DCASE & 8.70|11.46 & 7.78|10.39 & 6.79|9.27 & 5.75|\textbf{8.03} & 4.65|\textbf{6.54} & 6.73|\textbf{9.14} \\
    \rowcolor{lightgreen} & CDur & 16.25|18.36 & 11.12|\textbf{14.32} & 6.42|\textbf{10.40} & 3.11|6.55 & 1.45|3.93 & 7.67|10.71 \\
    \rowcolor{lightorange} & WSDDN & 26.56|7.17 & 18.74|4.23 & 12.31|2.19 & 6.62|1.13 & 2.80|0.48 & 13.41|3.05 \\
    \rowcolor{lightorange} & OICR & 25.88|6.12 & 16.79|3.50 & 9.87|1.75 & 4.95| 0.77 & 1.90|0.36 & 11.88|2.50 \\
    \rowcolor{lightorange} & PCL & 29.38|6.51 & 19.68|3.42 & 11.14|1.66 & 5.39|0.78 & 2.06|0.37 & 13.53|2.55 \\
    \rowcolor{lightred} & RSKP & \textbf{40.32}|\textbf{21.31} & \textbf{32.54}|9.74 & \textbf{23.27}|4.22 & \textbf{14.18}|1.80 & \textbf{5.54}|0.57 & \textbf{23.17}|7.53 \\
    \rowcolor{lightred} \multirow{-7}{*}{\rotatebox[origin=c]{90}{Opportunity}} 
    & CoLA & 8.27|10.05 & 5.33|5.77 & 2.82|2.83 & 1.34|1.26 & 0.65|0.53 & 3.68|4.09 \\
    \hhline{--------}
    
    \rowcolor{lightgreen} & DCASE & 9.90|14.49 & 6.84|11.54 & 4.53|9.52 & 2.54|7.42 & 1.57|5.43 & 5.08|9.68 \\
    \rowcolor{lightgreen} & CDur & 12.27|17.10 & 8.07|14.88 & 5.11|12.40 & 3.43|9.80 & 2.26|\textbf{8.82} & 6.23|12.60 \\
    \rowcolor{lightorange} & WSDDN & 13.02|10.44 & 9.58|8.48 & 6.70|6.45 & 4.75|4.43 & 2.94|2.59 & 7.40|6.48 \\
    \rowcolor{lightorange} & OICR & 13.21|26.61 & 9.05|19.55 & 5.26|13.51 & 2.95|7.40 & 1.23|3.75 & 8.46|14.17 \\
    \rowcolor{lightorange} & PCL & 23.11|\textbf{31.42}& 17.18|\textbf{22.61} & 9.85|\textbf{15.31} & 4.88|8.45 & 1.98|3.56 & 11.40|\textbf{16.27} \\
    \rowcolor{lightred} & RSKP & 15.71|23.00 & 10.89|17.71 & 6.62|14.53 & 3.92|\textbf{10.81} & 2.29|7.67 & 7.89|14.75 \\
    \rowcolor{lightred} \multirow{-7}{*}{\rotatebox[origin=c]{90}{WetLab}}
    & CoLA & \textbf{24.86}|14.24 & \textbf{20.93}|11.94 & \textbf{15.46}|10.24 & \textbf{11.35}|8.69 & \textbf{7.66}|6.48 & \textbf{16.05}|10.32 \\
    \hhline{--------}

    \rowcolor{lightgreen} & DCASE & 32.04|\textbf{44.77} & \textbf{25.36}|\textbf{37.66} & \textbf{17.58}|\textbf{27.88} & 9.88|17.11 & 4.04|8.28 & \textbf{17.78}|\textbf{27.14} \\
    \rowcolor{lightgreen} & CDur & 20.67|29.84 & 18.67|27.44 & 16.45|24.90 & \textbf{14.20}|\textbf{22.00} & \textbf{11.60}|\textbf{18.72} & 16.32|24.56 \\
    \rowcolor{lightorange} & WSDDN & 30.40|11.18 & 22.20|6.77 & 14.01|3.89 & 7.21|2.10 & 2.64|0.89 & 15.29|4.97 \\
    \rowcolor{lightorange} & OICR & \textbf{32.45}| 8.75 & 22.39|4.22 & 13.52|2.03 & 6.43|0.97 & 2.33|0.41 & 15.42|3.28 \\
    \rowcolor{lightorange} & PCL & 28.02|8.78 & 16.43|3.91 & 7.98|1.75 & 3.32|0.77 & 1.05|0.35 & 11.36|3.11 \\
    \rowcolor{lightred} & RSKP & 23.99|31.27 & 17.10|22.84 & 11.15|14.90 & 5.87|7.62 & 2.35|3.09 & 12.09|15.94 \\
    \rowcolor{lightred} \multirow{-7}{*}{\rotatebox[origin=c]{90}{Hang-Time}}
    & CoLA & 14.84|14.96 & 11.79|11.14 & 8.39|7.33 & 5.05|4.11 & 2.36|1.80 & 8.49|7.87 \\
    \hhline{--------}

    \rowcolor{lightgreen} & DCASE & \textbf{66.26}|\textbf{72.87} & \textbf{58.40}|\textbf{70.41} & 49.17|\textbf{66.94} & 42.00|\textbf{59.31} & 32.96|\textbf{55.84} & \textbf{49.76}|\textbf{65.07} \\
    \rowcolor{lightgreen} & CDur & 26.73|29.39 & 19.46|26.84 & 13.71|24.93 & 9.63 |21.73 & 6.39|20.56 & 15.19|24.69 \\
    \rowcolor{lightorange} & WSDDN & 40.39|58.16 & 29.48|46.33 & 21.54|35.14 & 14.41|25.66 & 9.28|15.78 & 23.02|36.22 \\
    \rowcolor{lightorange} & OICR & 9.33|55.66 & 7.06|39.36 & 6.52|28.23 & 6.30|19.28 & 4.84|13.68 & 6.81|31.24 \\
    \rowcolor{lightorange} & PCL & 1.40|55.58 & 8.38|36.42 & 7.54|24.42 & 7.19|16.35 & 6.35|10.91 & 7.97|28.73 \\ 
    \rowcolor{lightred} & RSKP & 63.27|64.69 & 57.75|59.77 & \textbf{49.27}|52.41 & \textbf{43.84}|48.17 & \textbf{33.82}|44.61 & 49.59|53.93 \\
    \rowcolor{lightred} \multirow{-7}{*}{\rotatebox[origin=c]{90}{RWHAR}}
    & CoLA & 31.61|22.49 & 25.36|20.30 & 16.74|18.26 & 11.77|17.06& 2.60|16.31 & 17.62|18.88 \\
    \hhline{--------}

    \rowcolor{lightgreen} & DCASE & 34.06|38.57 & 28.29|34.21 & 21.68|29.90 & 13.33|26.99 & 7.13|24.24 & 20.90|30.78 \\
    \rowcolor{lightgreen} & CDur & 19.55|21.43 & 14.64|18.01 & 10.21|14.57 & 6.90|13.35 & 4.04|11.22 & 11.07|15.72 \\
    \rowcolor{lightorange} & WSDDN & 49.56|42.34 & 42.14|35.23 & 33.69|27.42 & 25.44|20.71 & 17.26|14.39 & 41.13|28.02 \\
    \rowcolor{lightorange} & OICR & 15.53|47.06 & 8.66|35.18 & 5.42|23.83 & 2.61|15.59 & 1.00|8.48 & 6.61|26.03 \\
    \rowcolor{lightorange} & PCL & 26.75|57.38 & 16.75|43.72 & 9.52|29.39 & 3.90|18.82 & 1.36|9.01 & 11.65|31.66 \\
    \rowcolor{lightred} & RSKP & \textbf{69.64}|\textbf{71.24} & \textbf{61.29}|\textbf{67.21} & \textbf{52.15}|\textbf{63.65} & 25.00|\textbf{59.62} & 12.71|\textbf{54.71} & \textbf{44.16}|\textbf{63.29} \\
    \rowcolor{lightred} \multirow{-7}{*}{\rotatebox[origin=c]{90}{WEAR}}
    & CoLA & 37.49|26.64 & 35.34|26.19 & 33.13|24.99 & \textbf{28.28}|23.26 & \textbf{23.31}|20.81 & 31.51|24.38 \\
    \hhline{--------}

    \rowcolor{lightgreen} & DCASE & 40.12|33.68 & 31.84|27.02 & 15.69|14.84 & 3.87|5.68 & 2.09|3.90 & 18.72|17.02 \\
    \rowcolor{lightgreen} & CDur & 20.81|21.00 & 9.67|9.56 & 5.46|5.59 & 2.81|2.81 & 0.46|0.48 & 7.84|7.89 \\
    \rowcolor{lightorange} & WSDDN & 57.16|5.11 & \textbf{49.92}|5.08 & 38.98|3.89 & 30.96|2.99 & 19.48|2.16 & 39.30|3.85 \\
    \rowcolor{lightorange} & OICR & 50.72|22.87 & 49.85|22.09 & \textbf{45.93}|19.25 & \textbf{37.82}|14.16 & \textbf{25.31}|8.73 & \textbf{41.93}|17.42 \\
    \rowcolor{lightorange} & PCL & 46.71|20.72 & 43.40|19.19 & 39.05|13.36 & 30.38|9.20 & 6.55|16.56 & 34.89|13.80 \\
    \rowcolor{lightred} & RSKP & \textbf{59.46}|\textbf{51.69} & 47.69|\textbf{34.80} & 35.25|\textbf{23.82} & 20.82|\textbf{16.05} & 11.38|7.69 & 34.92|\textbf{26.81} \\
    \rowcolor{lightred} \multirow{-7}{*}{\rotatebox[origin=c]{90}{XRFV2}}
    & CoLA & 41.61|28.79 & 34.54|24.05 & 27.87|20.40 & 19.18|14.88 & 12.27|\textbf{10.04} & 27.10|19.63 \\
    \hline
  \end{tabular}
    \caption{\textit{Comparison between window-level input and full-sequence input. Each cell reports \textbf{window | full}. The table shows average results over tIoU thresholds (0.3:0.1:0.7). Best results per dataset are highlighted in \textbf{bold}.}}
  \label{tab:input_result}

\end{table*}

\begin{figure}[h]
    \centering
    \includegraphics[width=1\linewidth]{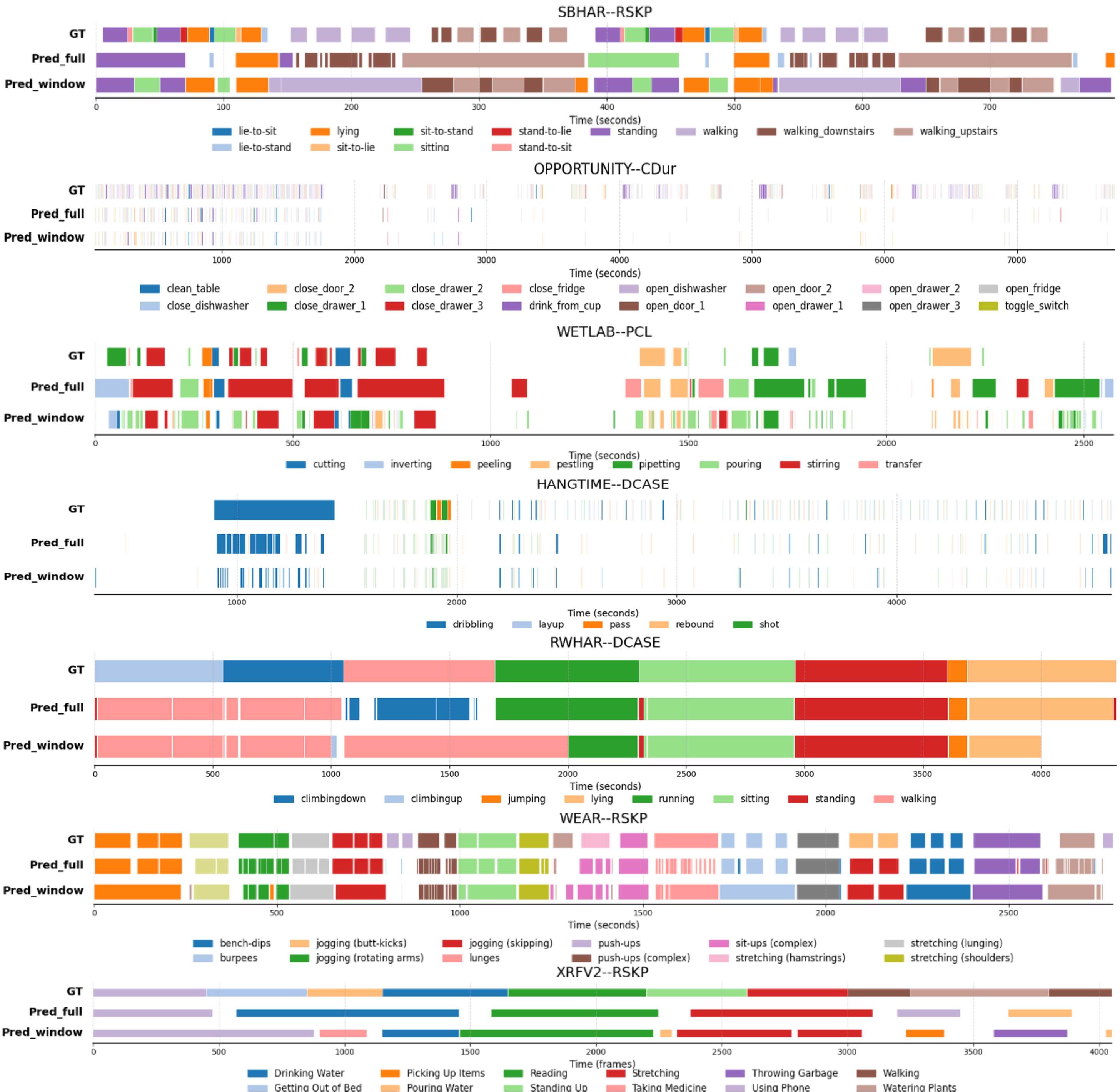}
    \caption{Visualization of predictions from the best-performing model for each dataset, comparing the \texttt{full\_input} (middle) and \texttt{window\_input} (bottom) inference strategies against the ground truth (top). Colors denote different action classes.}
    \label{fig:visualization}
\end{figure}

\emph{\textbf{(4) Visualization.}\label{sec:visualization}} To provide a more intuitive, qualitative analysis of model behavior, we identify the single \textbf{best-performing model for each dataset}. We then visualize the predictions of this specific model under both the \texttt{full\_input} and \texttt{window\_input} inference strategies. To generate clear visualizations from raw outputs, we apply a simple two-step post-processing: (i) for time-overlapping segments with different labels, keep only the one with the highest confidence; (ii) for segments with the same label that touch or overlap, merge them into a single segment.

The visualizations in Figure~\ref{fig:visualization} qualitatively affirm the strong performance of the RSKP model, which emerges as the top performer across several datasets. 
On the WEAR dataset, a clear pattern of ``over-segmentation" is evident. The model predicts a single, continuous ground truth action as numerous disconnected, short segments. This behavior is a direct manifestation of transferring video-based methods to the IMU domain, where the model tends to treat discrete ``IMU frames" as analogous to ``video frames," making localized, independent decisions that struggle to maintain long-range temporal continuity.
Conversely, the XRFV2 dataset, with its more uniform and continuous action segments, appears to better align with the model's operational sweet spot, resulting in predictions that are notably more coherent. Yet, this very continuity introduces an opposing failure mode: ``under-segmentation" and misclassification. As shown, the model incorrectly merges several distinct, consecutive ground truth actions into a single, monolithic predicted action.

These two contrasting examples from WEAR and XRFV2 reveal a core dilemma for the current model: it struggles to strike an ideal balance between capturing fine-grained action boundaries and maintaining long-range temporal consistency, often oscillating between over-segmentation and under-segmentation depending on the data's characteristics.

\emph{\textbf{(5) Impact of Subject Variability and Inference Strategy.}\label{sec:subject-variability-and-inference}}
A direct comparison between the intra-subject evaluation on the XRFV2 dataset (Table~\ref{tab:result_xrfv2_noCorss}) and the inter-subject LOSO results (Table~\ref{tab:result_supervised_weaklySupervised}) reveals a quantum leap in performance. The OICR model's mAP, for instance, skyrockets from 17.42\% in the cross-subject scenario to 67.88\% under the intra-subject, window-based inference. This vast improvement stems from the elimination of the ``domain gap" created by inter-subject variability. In a cross-subject setting, the model must learn a generic action representation that generalizes to all users. In the intra-subject setting, this challenge is removed, allowing the model to specialize in the personalized ``action fingerprint" of a single user, a substantially less complex task.

Within the intra-subject evaluation itself, the choice of inference strategy \texttt{window\_input} versus \texttt{full\_input} also proves to be a critical factor. The \texttt{window\_input} strategy consistently outperforms the \texttt{full\_input} approach, particularly for models originating from the image domain like WSDDN, OICR, and PCL. The WSDDN model, for example, achieves a 55.88\% mAP with windowed input, which then plummets to just 11.20\% when inferring on the full sequence. This occurs because partitioning a long sequence into shorter windows effectively simplifies the problem. 
For fragile architectures that rely on random sampling for proposal generation, searching within a short window drastically reduces noise and interference from irrelevant data, thereby increasing the probability of generating high-quality proposals. Conversely, performing inference on the entire long sequence at once causes these proposal mechanisms to get lost in an overly large search space.

\begin{table*}[h]
  \centering
  \footnotesize

  \begin{tabular}{@{}llccc|cccccc|c@{}}
    \hline
    & Model & P ($\uparrow$) & R ($\uparrow$) & F1 ($\uparrow$) & UR ($\downarrow$) & OR ($\downarrow$) & DR ($\downarrow$) & IR ($\downarrow$) & FR ($\downarrow$) & MR ($\downarrow$) & mAP ($\uparrow$) \\ 
    \hline

    
    \rowcolor{lightgreen} & DCASE & 27.94|28.63 & 20.62|20.61 & 21.37|21.44 & 0.71|0.68 & 32.05|29.41 & 1.89|1.93 & 25.20|27.20& 0.01|0.02 & 0.54|0.46 & 9.04 | 8.02 \\
    \rowcolor{lightgreen} & CDur & 21.45|21.83 & 15.82|16.21 & 16.33|16.77 & \textbf{0.44}|0.45 & 32.16|31.62 & 2.12|2.10 & 21.25|24.79& 0.01|0.01 & 1.09|1.01 & 5.20|6.67 \\
    
    \rowcolor{lightorange} & WSDDN & 62.92|43.29 & 63.33|20.10 & 62.71|25.90 & 0.62|\textbf{0.10} & 17.99|37.86 & 0.40|2.58 & 3.65|\textbf{1.74} & 0.01|0.00 & \textbf{0.00}|0.11 & 55.88|11.20 \\
    \rowcolor{lightorange} & OICR & \textbf{64.22}|51.45 & \textbf{65.84}|\textbf{53.00} & \textbf{64.65}|\textbf{51.30} & 0.55|0.65 & \textbf{17.06}|30.44 & \textbf{0.36}|\textbf{0.80} & 3.22|2.95 & \textbf{0.00}|\textbf{0.00} & 0.03|0.11 & \textbf{67.88}|\textbf{45.81}\\
    \rowcolor{lightorange} & PCL & 61.89|50.31 & 63.21|51.49 & 62.10|50.32 & 0.64|0.70 & 20.37|29.18 & 0.39|0.81 & \textbf{2.32}|5.40 & \textbf{0.00}|\textbf{0.00} & 0.02|0.10 & 64.30|43.74 \\
    
    \rowcolor{lightred} & RSKP & 49.48|40.59 & 39.91|31.70 & 40.43|32.98 & 0.71|0.41 & 22.32|31.68 & 0.86|1.64 & 9.55|8.72 & 0.06|0.01 & 0.32|0.98 & 32.99 | 21.24 \\
    
    \rowcolor{lightred} \multirow{-7}{*}{\rotatebox[origin=c]{90}{XRFV2}} 
    & CoLA & 48.18|\textbf{51.78}& 37.40|28.21& 39.84|33.38& 0.52|0.38 & 19.38|\textbf{12.13} & 1.23|1.60 & 20.57|25.47 & 0.05|0.09 & 0.26|\textbf{0.01} & 21.15 | 16.67\\
    
    \hline
  \end{tabular}
    \caption{\textit{Intra-subject evaluation of weakly supervised temporal action localization on the XRFV2 dataset:} The table provides per-sample classification metrics, i.e. Precision (P), Recall (R), F1-Score (F1), misalignment ratios \cite{ward2006evaluating} and average mAP applied at different tIoU thresholds (0.3:0.1:0.7). Each cell reports \textbf{window | full}.  Best results per dataset are in \textbf{bold}.}
  \label{tab:result_xrfv2_noCorss}
  
\end{table*}

\vspace{-20pt}

\section{Discussion}\label{sec:discussion}

This work shows that weakly supervised learning is a viable and promising paradigm for temporal action localization in IMU data. However, it also reveals a fundamental tension: while models can learn from only sequence-level labels, their performance is constrained by the unique characteristics of inertial signals. This discussion explores these challenges and charts a path forward, thereby directly answering \textit{\textbf{RQ3 (Insights \& Future Directions)}} by summarizing dominant failure modes and outlining concrete opportunities for advancing WS-IMU-TAL.

\paragraph{\textbf{(1) The Unique Challenges of Weak Supervision on IMU Data.}}
A key takeaway is that simply porting MIL pipelines from other domains is insufficient, as inertial signals present unique challenges in feature representation, boundary definition, and action composition that are absent in other modalities.

\textbf{Challenges from Dataset Heterogeneity.} 
Compared to \textbf{audio}, which is also a 1D signal, sound analysis often leverages high-dimensional spectrograms with rich frequency channels. Models can afford to downsample across this channel dimension to learn robust features. In contrast, many IMU datasets are low-dimensional, sometimes comprising only three accelerometer axes. For such data, any aggressive channel-wise downsampling can lead to a catastrophic loss of information, making feature extraction more brittle.
Compared to \textbf{images}, where object boundaries are often defined by rich visual cues like color, texture, and edges, IMU data lacks such static, descriptive features. An action boundary in an IMU stream is not a visual edge but a \textit{temporal event}—often characterized by an abrupt change in signal dynamics. This makes it challenging for standard MIL models, which excel at finding discriminative static features, to distinguish a subtle but meaningful temporal transition from simple signal noise.
Furthermore, unlike \textbf{videos}, which can visually represent multiple, concurrent actions with distinct spatial locations, an IMU sensor inherently entangles all motions of a body part into a single, composite signal. For example, the action of "eating while walking" is visually separable in a video—one can see the hand moving to the mouth and the legs moving independently. However, a wrist-worn IMU would record a complex superposition of both the large-scale motion from walking and the fine-grained motion from eating. Disentangling such a secondary, subtle action from the dominant primary action is nearly an ill-posed problem for a single sensor under a weak supervision paradigm.

\textbf{Potential Solutions.} These challenges necessitate IMU-specific solutions. To counteract feature sparsity, one could employ targeted feature engineering, augmenting the raw signal with derivatives, frequency-domain features (e.g., FFT), or wavelet transforms before feeding them to the model. To better capture temporal event boundaries, models could be designed to be explicitly sensitive to signal change-points, perhaps by using the temporal derivative of the signal as an additional input channel. Finally, to address action entanglement, multi-scale architectures could be explored. Such models would analyze the signal at different temporal resolutions simultaneously, allowing one branch to focus on the low-frequency, dominant action (walking) while another focuses on the high-frequency, subtle action (eating), enabling better separation and localization.

\paragraph{\textbf{(2) The Bottleneck of Temporal Proposal Generation.}}

Our methodology's reliance on a two-stage pipeline exposes a major bottleneck: the generation of high-quality temporal proposals.

\textbf{Challenges from Proposals.} Two-stage WSOD methods in computer vision are successful in large part due to sophisticated, class-agnostic algorithms like Selective Search~\cite{van2011segmentation}, which intelligently generate high-quality candidate regions before any classification occurs. The IMU domain lacks an equivalent. Our current approach, using a simple stochastic sampling strategy, is content-agnostic and cannot guarantee optimal coverage of potential action segments. This creates a hard ceiling on final performance: an action can never be localized, no matter how powerful the classifier, if a good temporal proposal covering it was never generated in the first place.

\textbf{Potential Solutions.} Future research should focus on developing dedicated, content-aware temporal proposal algorithms specifically for IMU data. One promising direction is to leverage classical signal processing techniques. For example, algorithms based on unsupervised change-point detection could automatically identify moments of significant statistical shift in the signal, which often correlate with the start or end of an action. Another approach is to use a lightweight, pre-trained action model. This model could be trained on a diverse set of data to provide a score for each time step indicating the likelihood that an action is occurring, with high-scoring contiguous regions forming the basis for proposals. Both approaches would move beyond random sampling to intelligently guide the localization process, breaking the performance ceiling imposed by our current method.

\paragraph{\textbf{(3) The Critical Need for a Unified Pretraining Foundation.}}

A significant bottleneck in the IMU research community is the lack of a large-scale, unified pretraining foundation analogous to ImageNet~\cite{deng2009imagenet}. While recent works~\cite{wei2025one,saha2025pulse} have begun to pioneer foundational models for inertial data, building a broadly transferable IMU foundation model remains challenging.

\textbf{Challenges from Pretrained Models.}
\textit{Data heterogeneity} is only the first obstacle: IMU signals exhibit high intra-class, inter-subject variability (e.g., ``running" differs substantially across age, style, and physiology), and this variability is further amplified by hardware differences (sensor placement, orientation, sampling rate, sensor noise, and device-specific filtering).
Beyond data collection, there are additional training-side challenges.
First, \textit{representation alignment} across datasets is non-trivial: different coordinate frames, missing modalities (ACC-only vs. ACC+GYRO), and inconsistent preprocessing pipelines often make ``the same" channel semantically different.
Second, \textit{pretraining objective mismatch} is common: generic self-supervised learning objectives may learn invariances that are useful for classification but blur fine-grained temporal boundaries that are critical for localization.
Third, \textit{evaluation fragmentation} slows progress: without a standardized pretraining benchmark (datasets, protocols, and downstream suites), it is difficult to identify which ingredients (architecture, objective, augmentation, or data mixture) truly improve transfer.

\textbf{Potential Solutions.}
A practical path forward should combine data standardization, better pretraining recipes, and data scaling.
1) \textit{Standardize and normalize inputs:} establish community-wide conventions for sensor placement and metadata (sampling rate, coordinate frame, and initial orientation), and promote canonical transformations (e.g., gravity alignment / orientation quaternion outputs) to reduce cross-dataset ambiguity.
2) \textit{Develop IMU-specific foundation training objectives:} beyond standard contrastive learning, explore objectives that preserve temporal structure (e.g., masked signal modeling with boundary-aware masking, multi-resolution prediction, or change-point-aware auxiliary losses) so that representations transfer not only to recognition but also to localization.
3) \textit{Train for heterogeneity explicitly:} use multi-dataset pretraining with lightweight domain adapters (or normalization layers) to handle device- and dataset-specific shifts while keeping a shared backbone.
4) \textit{Scale data via synthesis:} complement real-world corpora with physics/biomechanics-inspired simulation and generative augmentation (e.g., trajectory perturbations, realistic noise injection, or generative models that respect sensor constraints) to expand coverage of rare actions and sensor configurations.

Together, these directions can move the community beyond ad-hoc dataset-specific pretraining toward a truly general-purpose IMU foundation model.

\section{Conclusion}\label{sec:conclusion}

\benchName provides a reproducible benchmark for weakly supervised IMU temporal action localization.
Across seven public datasets, we evaluate representative methods transferred from audio, image, and video, and identify consistent trends: transfer is modality-dependent, weak supervision is most effective on datasets with longer actions and richer sensing, and major bottlenecks stem from short actions, temporal ambiguity, and proposal quality. Beyond reporting numbers, \benchName provides actionable insights into failure modes and highlights key directions for future WS-IMU-TAL research, including IMU-specific temporal proposal generation, boundary-aware learning objectives, and stronger temporal reasoning under weak supervision.
We hope this benchmark, together with its protocols and analyses, will serve as a common foundation for the community to build scalable WS-IMU-TAL systems without prohibitive annotation costs.

\bibliographystyle{ACM-Reference-Format}
\bibliography{imwut}


\end{document}